\documentclass[journal]{IEEEtran}
\ifCLASSINFOpdf
\else
\fi
\usepackage[cmex10]{amsmath}
\usepackage{graphicx,subfigure}
\graphicspath{ {./figs/} }

\usepackage{amssymb}
\usepackage{amsthm}
\usepackage{amsmath}
\usepackage{algorithm}               
\usepackage{algorithmic}             
\usepackage{url}
\usepackage{color}

\usepackage{graphicx}

\usepackage{algorithmic}

\usepackage{array}
\usepackage{url}

\usepackage{color}

\usepackage{graphicx}

\newcommand{\mcb}{\color{black}}

\def\cs3c{C$\text{S}^3$C}
\def\s3c{$\text{S}^3$C}
\def\ie{i.e.}
\def\eg{e.g.}

\def\etal{\textit{et al.}}
\def\st{\textrm{s.t.}}
\def\diag{\textrm{diag}}

\def\rank{\textrm{rank}}

\def\trace{\textrm{trace}}

\def\a{\boldsymbol{a}}
\def\c{\textbf{c}}
\def\q{\textbf{q}}

\def\0{\textbf{0}}
\def\1{\textbf{1}}
\def\x{\boldsymbol{x}}
\def\I{\mathbf{I}}

\def\C{\mathcal{C}}

\def\E{\mathcal{E}}

\def\Q{\mathcal{Q}}

\def\S{\mathcal{S}}

\newcommand{\RR}{I\!\!R} 

\newcommand{\myparagraph}[1]{\smallskip\noindent\textbf{#1.}}

\DeclareMathOperator*{\argmin}{arg\,min}

\newtheorem{problem}{Problem}[]

\begin{document}
%
\title{Structured Sparse Subspace Clustering: A Joint Affinity Learning and Subspace Clustering Framework}
%
%
%

\author{Chun-Guang~Li,~Chong~You,~and~Ren\'{e}~Vidal
\thanks{C.-G. Li is with the School of Information and Communication Engineering, the Beijing University of Posts and Telecommunications, Beijing 100876, P.R. China. E-mail: lichunguang@bupt.edu.cn.}
\thanks{C. You and R. Vidal are with the Center for Imaging Science, the Johns Hopkins University, Baltimore, MD 21218, USA. E-mail: \{cyou, rvidal\}@cis.jhu.edu.}
}

\maketitle

\begin{abstract}
Subspace clustering refers to the problem of segmenting data drawn from a union of subspaces. State-of-the-art approaches for solving this problem follow a two-stage approach. In the first step, an affinity matrix is learned from the data using sparse or low-rank minimization techniques. In the second step, the segmentation is found by applying spectral clustering to this affinity. While this approach has led to state-of-the-art results in many applications, it is sub-optimal because it does not exploit the fact that the affinity and the segmentation depend on each other. In this paper, we propose a joint optimization framework --- Structured Sparse Subspace Clustering (\s3c) --- for learning both the affinity and the segmentation. The proposed \s3c framework is based on expressing each data point as a structured sparse linear combination of all other data points, where the structure is induced by a norm that depends on the unknown segmentation. Moreover, we extend the proposed \s3c framework into Constrained Structured Sparse Subspace Clustering (\cs3c) in which available partial side-information is incorporated into the stage of learning the affinity. We show that both the structured sparse representation and the segmentation can be found via a combination of an alternating direction method of multipliers with spectral clustering. Experiments on a synthetic data set, the Extended Yale B face data set, the Hopkins 155 motion segmentation database, and three cancer data sets demonstrate the effectiveness of our approach.
\end{abstract}


%
\IEEEpeerreviewmaketitle


\section{Introduction}

In many real-world applications, we need to deal with high-dimensional datasets, such as images, videos, text, and more. In practice,  such high-dimensional datasets can often be well approximated by multiple low-dimensional subspaces corresponding to multiple classes or categories. For example, the feature point trajectories associated with a rigidly moving object in a video lie in an affine subspace (of dimension up to 3) \cite{Tomasi:IJCV92}, and face images of a subject under varying illumination lie in a linear subspace (of dimension up to 9) \cite{Ho:CVPR03}. Therefore, the task, known in the literature as \emph{subspace clustering}, is to segment the data into their corresponding subspaces. This problem has many applications, \eg, image representation and compression \cite{Hong:TIP06}, motion segmentation \cite{Costeira:IJCV98, Rao:PAMI10}, and temporal video segmentation \cite{Vidal:PAMI05} in computer vision; hybrid system identification in control \cite{Bako:Automatica11}; community clustering in social networks \cite{Jalali:ICML11}; and genes expression profiles clustering in bioinformatics \cite{McWilliams:DMKD14}.

\myparagraph{Previous Work}
The subspace clustering problem has received a lot of attention over the past few years and many methods have been developed, including iterative methods \cite{Bradley:JGO00, Tseng:JOTA00, Zhang:WSM09, Agarwal:ACM04}, algebraic methods \cite{Vidal:PAMI05,Ma:SIAM08,Huang:CVPR04-ED,Tsakiris:AffineASC-ArXiv15}, statistical methods \cite{Boult:WMU91,Leonardis:PR02,Archambeau:Neuro08,Gruber-Weiss:CVPR04,Ma:PAMI07,Yang:CVPR06}, and spectral clustering based methods \cite{Yan:ECCV06,Goh:CVPR07,Fan:PAMI06,Chen:IJCV09,Zhang:IJCV12,Elhamifar:CVPR09,Elhamifar:TPAMI13,Liu:ICML10,Liu:TPAMI13,Favaro:CVPR11,Vidal:PRL14,Lu:ECCV12,Lu:ICCV13-TraceLasso,
Lu:ICCV13,Wang:NIPS13-LRR+SSC,Dyer:JMLR13,Heckel:TIT15,Zhang:ICCV13,Park:NIPS14,Li:CVPR15MoG,You:CVPR16-EnSC} (see \cite{Vidal:SPM11-SC} for details). Among them, methods based on spectral clustering have become extremely popular. Such methods divide the problem in two steps. In the first one, an affinity matrix is learned from the data. In the second one, spectral clustering is applied to this affinity. Arguably, the first step is the most important, as the success of the spectral clustering algorithm is largely dependent on constructing an informative affinity matrix.

Recent methods for learning the affinity matrix are based on the self-expressiveness model, which states that a point in a union of subspaces can be expressed as a linear combination of other data points, \ie, $X = XC$, where $X \in \RR^{D \times N}$ is the data matrix containing $N$ data points as its columns, and $C \in \RR^{N\times N}$ is the coefficients matrix, also known as the \emph{representation matrix}. With corrupted data, this constraint is relaxed to $X = XC + E$, where $E$ is a matrix of errors. The self-expressiveness model is then formulated as the following optimization problem:
\begin{equation}
\min _{C,E} \| C \|_\C  + \lambda \|E\|_{\E} ~~\text{s.t.}~~ X = XC + E, ~~ \diag(C) = 0,
\label{eq:self-expression}
\end{equation}
where $\|\cdot\|_\C$ and $\|\cdot\|_\E$ are two properly chosen norms, $\lambda > 0$ is a tradeoff parameter, and the constraint $\diag(C) = 0$ is optionally used when, \eg, $\| C \|_\C = \|C\|_1$, to rule out the trivial solution of $C$ being an identity matrix.


The primary difference between different methods lies in the choice of regularization on $C$ and/or the noise term $E$. Currently, the $\ell_0$ norm, $\ell_1$ norm, $\ell_2$ norm or the Frobenius norm, the nuclear norm ($\|\cdot\|_\ast$) \cite{Fazel2002}, the TraceLasso norm \cite{Grave:NIPS11}, and mixtures of some of them\footnote{The relations among the nuclear norm based models and the connections between the nuclear norm and the Frobenius norm are discussed in \cite{Zhang:NeuralCom15,Peng:TNNLS16}.}, \eg, $\ell_1+\ell_2$, $\ell_1+ \|\cdot\|_\ast$, $\ell_{2,1}$, have been well exploited as the choices of regularization. For example,
in Sparse Subspace Clustering (SSC) \cite{Elhamifar:CVPR09,Elhamifar:TPAMI13,Elhamifar:ICASSP10,You:CVPR16-SSCOMP}, the $\ell_1$ norm is used for $\|\cdot\|_\C$ as a convex surrogate over the $\ell_0$ norm to promote sparseness in $C$, and the Frobenius norm and/or the $\ell_1$ norm of $E$ is used to handle Gaussian noise and/or outlying entries;
in Low-Rank Representation (LRR) \cite{Liu:ICML10,Liu:TPAMI13},  Low-Rank Subspace Clustering (LRSC) \cite{Favaro:CVPR11,Vidal:PRL14}, and Multiple Subspace Recovery (MSR) \cite{Luo:ECML11}, the nuclear norm $\|\cdot\|_\ast$ is adopted for $\|\cdot\|_\C$ as a convex surrogate of the $\text{rank}$ function, the $\ell_{2,1}$ norm of $E$ is used to tackle outliers (in LRR), the $\ell_1$ norm of $E$ is used to handle outlying entries (in MSR), and the Frobenius norm and/or the $\ell_1$ norm of $E$ is used to handle Gaussian noise and/or outlying entries (in LRSC);
in Least Squares Regression (LSR) \cite{Lu:ECCV12} and $\ell_2$-graph\footnote{
The recipe of using only dominant coefficients has also been exploited in manifold learning \cite{Li:ACCV09,Elhamifar:NIPS11}.}
\cite{Peng:TCYB16}, the Frobenius norm is used for $\|\cdot\|_\C$ and $\|\cdot\|_\E$ to regularize the representation matrix $C$ and to handle the Gaussian noise $E$;
in Correlation Adaptive Subspace Segmentation (CASS) \cite{Lu:ICCV13-TraceLasso}, Low-Rank Sparse Subspace Clustering (LRSSC) \cite{Wang:NIPS13-LRR+SSC}, and Elastic Net Subspace Clustering (EnSC) \cite{You:CVPR16-EnSC}, the TraceLasso norm, the mixture of $\ell_1+ \|\cdot\|_\ast$, and the mixture of $\ell_1+\ell_2$ norm are employed as $\|\cdot\|_\C$, respectively, to gain a balance between connectivity and correctness;
in Thresholding Subspace Clustering (TSC) \cite{Heckel:TIT15}, SSC by Orthogonal Matching Pursuit (SSC-OMP)~\cite{Dyer:JMLR13,You:CVPR16-SSCOMP,Tschannen:arXiv16}, $\ell_0$-SSC \cite{Wang:AISTAT16,Yang:ECCV16}, and Nearest Subspace Neighbor (NSN) \cite{Park:NIPS14}, the $\ell_0$ norm is investigated.
%
In addition, weighting $C$ with locality or spatial information \cite{Pham:CVPR12, Hu:CVPR14}, and computing $C$ in latent feature space \cite{Patel:ICCV13,Patel:ICIP14,Patel:JSTSP15} or with a learnt dictionary~\cite{Shen:ICML16} have also been proposed.
%

Once the coefficients matrix $C$ is found by any of the methods above, the segmentation of the data can be obtained by spectral clustering~\cite{vonLuxburg:StatComp2007}, which computes an embedding of the data from the affinity matrix induced from $C$, \eg, $|C|+|C^\top|$, and then obtains the segmentation of the data by $k$-means.

While the above approaches have been incredibly successful in many applications, an important disadvantage is that they divide the problem into two separate stages: affinity learning (using, e.g., SSC, LRR, LRSC, LSR) and spectral clustering. Dividing the problem in two steps is, on the one hand, appealing because the first step can usually be solved using convex optimization techniques, while the second step can be solved using existing spectral clustering techniques. On the other hand, its major disadvantage is that the natural relationship between the affinity matrix and the segmentation of the data is not explicitly captured.

In this paper, we attempt to integrate these two separate stages into one unified optimization framework. One important observation is that a perfect subspace clustering can often be obtained from an imperfect affinity matrix. In other words, the spectral clustering step can correct errors in the affinity matrix, which can be viewed as a process of information gain by denoising. Because of this, if we feed back the information gain properly, it may help the self-expressiveness model yield a better affinity matrix. As shown in Fig.~\ref{Fig:Z-Theta-visualization-3}, the clustering results can help the self-expressiveness model find an improved affinity matrix and thus boost the final clustering results.

\myparagraph{Paper Contributions}
In this paper, we propose a new approach to subspace clustering called Structured Sparse Subspace Clustering (SSSC or \s3c), which integrates the two separate stages of computing a sparse representation matrix and applying spectral clustering into a unified optimization framework. The proposed approach is based on minimizing a new subspace structured $\ell_1$ norm, which augments the $\ell_1$ norm with a segmentation dependent term. The resulting optimization problem is solved in an alternating minimization framework where the output of spectral clustering is used to define a subspace segmentation matrix, which is then used to re-weight the representation matrix in the next iteration. Depending on how the subspace segmentation matrix is defined, we obtain two different implementations of the \s3c framework:
\begin{itemize}
\item \textit{Hard} \s3c: in this case the segmentation produced by spectral clustering (\ie, after the $k$-means step) is used to construct a binary segmentation matrix for re-weighting the representation matrix in the next iteration. {\mcb This approach is studied in a preliminary version~\cite{Li:CVPR15} of our work.

\item \textit{Soft} \s3c: in this case the embedding of data produced by spectral clustering (\ie, before the $k$-means step) is used to construct a continuous real-valued segmentation matrix for re-weighting the representation matrix in the next iteration. This extension not only leads to a more principled optimization framework, but also has better empirical performance as it captures more information from the previous iteration.   }

\end{itemize}
{\mcb In addition, we extend the \s3c framework into a Constrained \s3c (\cs3c) framework, which enables us to perform subspace clustering with the help of partial pairwise side-information (e.g., prior knowledge about which points belong to the same group and which points do not belong to the same group).} Finally, we demonstrate the effectiveness of the proposed approach on a synthetic data set, the Extended Yale B face data set, the Hopkins 155 motion segmentation database, and three cancer data sets. Experimental results also show that, with the help of some side-information, the clustering accuracy could be improved.
{\mcb Compared to our preliminary work \cite{Li:CVPR15}, more experimental evaluations and discussions are presented.}


\myparagraph{Paper Outline}
The remainder of this paper is organized as follows. Section~\ref{sec:unified-optimization-framework} describes our unified optimization framework.
Section~\ref{sec:solving-algorithm} proposes algorithms for solving the problem.
Section~\ref{sec:discussions} presents some justification and discussions.
Section~\ref{sec:experiments} shows experiments and Section~\ref{sec:conclusion} presents the conclusions.

\section{A Unified Optimization Framework for Subspace Clustering}
\label{sec:unified-optimization-framework}

In this paper we address 
the following problem.

\begin{problem}[\bf Subspace clustering]
\label{pro:problem}
Let $X \in \RR^{D \times N}$ be a real-valued matrix whose columns are drawn from a union of $n$ subspaces of $\RR^D$, $\bigcup_{j=1}^n \{S_j\}$, of dimensions $d_j \ll \min \{D, N\}$, for $j=1,\dots,n$. The goal of subspace clustering is to segment the columns of $X$ into their corresponding subspaces.
\end{problem}

To begin with, we introduce some additional notations. Let $Q = \begin{bmatrix}\textbf{q}_1,\cdots,\textbf{q}_n \end{bmatrix}$ be an $N\times n$ binary matrix indicating the membership of each data point to each subspace. 
That is, ${q}_{ij}=1$ if the $i$-th column of $X$ lies in subspace $S_j$ and $q_{ij} = 0$ otherwise. We assume that each data point lies in only one subspace, hence if $Q$ is a valid segmentation matrix, we must have $Q \1= \1$, where $\1$ is the vector of all ones of appropriate dimension. Note that the number of subspaces is equal to $n$, so we must have that $\rank(Q) = n$. Thus, 
the space of all valid segmentation matrices with $n$ groups is:
\begin{align}
\Q = \{ Q \in \{0,1\}^{N\times n} : Q\1 = \1 ~ \text{and} ~ \rank(Q) = n \}.
\end{align}

\subsection{Structured Subspace Clustering: A Unified Framework}
\label{sec:st-sc}

Recall from \eqref{eq:self-expression} that data in a union of subspaces are self-expressive, that is, each data point in a union of subspaces can be expressed as a linear combination of other data points as $X = XC$ with $\diag(C)=\0$, where $C\in\RR^{N\times N}$ is coefficients matrix whose $(i,j)$ entry $C_{ij}$ captures the similarity between points $i$ and $j$. For the purpose of data clustering, we expect the coefficients matrix to be \textit{subspace-preserving} \cite{Vidal:Springer16}, \ie, $C_{ij} \neq 0$ only if points $i$ and $j$ lie in the same subspace. In existing approaches~\cite{Elhamifar:CVPR09, Liu:ICML10, Lu:ECCV12}, one searches for a subspace-preserving representation by adding some regularization on $C$ (\eg, $\ell_1$~\cite{Elhamifar:CVPR09}, $\|\cdot \|_\ast$~\cite{Liu:ICML10}, $\|\cdot\|_F$~\cite{Lu:ECCV12}) and solving the optimization program in \eqref{eq:self-expression}.

Once the representation matrix $C$ is computed, one defines the data affinity matrix as
\begin{equation}
	A = \frac{1}{2}(|C|+|C^\top|).
	\label{eq:def-Affinity}
\end{equation}
Notice that the data affinity matrix $A$ encodes the pairwise similarity and can also be interpreted as a cost matrix, in which each entry $A_{ij}$ specifies a cost for segmenting data points $\x_i$ and $\x_j$ into two different clusters. Given the affinity $A$, the clustering of the data is obtained by finding a segmentation matrix $Q$ that minimizes the sum of such costs, i.e.,
\begin{align}
	\min_{Q}  \frac{1}{2}\sum_{i,j} A_{i,j} \|\q^{(i)} - \q^{(j)}\|^2_2 \quad \st \quad Q \in \Q,
	\label{eq:cut}
\end{align}
where $\q^{(i)}$ and $\q^{(j)}$ are the $i$-th and $j$-th row of matrix $Q$, respectively. In practice, since the search over all $Q \in \Q$ is combinatorial, spectral clustering techniques \cite{vonLuxburg:StatComp2007} usually relax the constraint $Q \in \Q$ to $Q^\top Q = I$ and apply $k$-means to the rows of $Q$ to get the binary segmentation matrix.

The common framework for subspace clustering adopted by previous approaches, \eg,~\cite{Elhamifar:TPAMI13, Liu:ICML10, Lu:ECCV12}, divides the procedure into two independent steps: a) compute the representation matrix $C$, and then b) apply spectral clustering to the affinity matrix $A$. Unfortunately, it fails to exploit the correlations between the two steps.

Notice that both the representation matrix $C$ and the segmentation matrix $Q$ try to capture the segmentation of the data. {\mcb To quantify the interaction between matrix $C$ and $Q$, we propose a notion called \textit{subspace structured norm} of representation matrix $C$ with respect to $Q$~\cite{Li:CVPR15}, \ie,
\begin{align}
\label{eq:structured-Q-norm}
\begin{split}
\|C\|_Q &\doteq\sum_{i,j}|C_{ij}|(\frac{1}{2}\|\q^{(i)} - \q^{(j)}\|^2).
\end{split}
\end{align}
The $\|C\|_Q$ measures the disagreement between representation matrix $C$ and segmentation matrix $Q$. Suppose $Q$ was correct, then $\|C\|_Q$ vanishes if $C$ is subspace-preserving; otherwise, it is positive.\footnote{Strictly, $\|C\|_Q$ is not a norm but a semi-norm of $C$ for a given $Q$.}} On the other side, by substituting the affinity $A_{ij}=\frac{1}{2}(|C_{ij}|+|C_{ji}|)$ defined in \eqref{eq:def-Affinity} into $\|C\|_Q$, we identify the equivalence between the subspace structured norm $\|C\|_Q$ and the objective for spectral clustering in~\eqref{eq:cut}, \ie,
\begin{align}
\label{eq:structured-Q-norm}
\begin{split}
\|C\|_Q =\frac{1}{2}\sum_{i,j}A_{i,j} \|\q^{(i)} - \q^{(j)}\|^2_2,
\end{split}
\end{align}
which is thus established the connection between $\|C\|_Q$ and spectral clustering.

Now, we are ready to derive a joint optimization framework for subspace clustering in a natural 
way. Notice that the goal in the optimization program \eqref{eq:cut} is to search for a matrix $Q$ that minimizes the segmentation cost with $A$ fixed, and also that $A$ is defined from representation matrix $C$, which in turn is obtained by searching over all possible representation matrices that minimize the regularization terms in the optimization program \eqref{eq:self-expression}.
Therefore, we can combine the two optimization programs and search for $Q$ and $C$ at the same time in a joint optimization framework. Precisely, we derive a joint optimization framework for subspace clustering as follows:
\begin{align}
\label{eq:STSC-outliers}
\begin{split}
\min\limits_{C,E,Q} ~&
\alpha\|C\|_Q + \|C\|_{\C} + \lambda \|E\|_{\E} \\
\st ~~ &X = X C + E, ~\text{diag}(C) = \0, ~~ Q \in \Q,
\end{split}
\end{align}
where $\alpha >0$ is a trade-off parameter. In this framework, we search jointly for the representation matrix $C$ (and $E$), which satisfies the data self-expressiveness model $X=XC+E, \text{diag}(C) = \0$, and the data segmentation matrix $Q \in \Q$. The regularization term $\|C\|_\C$ is as before to induce the optimal solution for $C$ to have the subspace-preserving property. The term $\|C\|_Q$ is used to induce a representation matrix $C$ that minimizes the inter-cluster affinity. {\mcb Comparing to existing approaches as formulated in~\eqref{eq:self-expression}, we call the joint optimization framework \eqref{eq:STSC-outliers} as \textit{structured subspace clustering}.\footnote{Comparing to \cite{Li:CVPR15}, the unified
framework introduced here 
includes an additional term $\|C\|_\C$ in the objective and thus is more practical.}

}

The optimization problem in \eqref{eq:STSC-outliers} can be solved by alternatingly solving $(C, E)$ and $Q$. Given $Q$, the problem is a convex program for $(C, E)$, which can be solved efficiently using the alternating direction method of multipliers (ADMM) \cite{Boyd:FTML10, Lin:09}. On the other hand, given $(C, E)$, the solution for $Q$ can be computed approximately by spectral clustering.

\subsection{Structured Sparse Subspace Clustering (\s3c)}
\label{sec:st-ssc}

Among the many norms used by existing  subspace clustering algorithms, \eg, the $\ell_1$ norm used by SSC, the nuclear norm used by LRR and LRSC, and the Frobenius norm used by LSR, we adopt the $\ell_1$ norm in our structured subspace clustering framework in \eqref{eq:STSC-outliers},
and combine $\|C\|_1$ with subspace structured norm $\|C\|_Q$ to define a \textit{subspace structured} $\ell_1$ norm of $C$ as
\begin{align}
\label{eq:subspace-structured-L-1-Q}
\begin{split}
\|C\|_{1,Q} &\doteq  \| C \|_1 + \alpha \|C\|_Q \\
&= \sum_{i,j}|C_{ij}|(1 + \frac{\alpha}{2}\|\q^{(i)} - \q^{(j)}\|^2),
\end{split}
\end{align}
where $\alpha > 0$ is a tradeoff parameter. Clearly, the first term is the standard $\ell_1$ norm used in SSC. Therefore the subspace structured $\ell_1$ norm can be viewed as an $\ell_1$ norm augmented by an extra penalty on $C_{ij}$ when points $i$ and $j$ are in different subspaces according to the segmentation matrix $Q$. {\mcb By doing so, the segmentation information in $Q$ is incorporated in finding of subspace-preserving solution.} The reasons we prefer to use the $\ell_1$ norm are two-fold:
\begin{itemize}
\item Since both the $\ell_1$ norm and the subspace structured norm are based on the $\ell_1$ norm, this leads to a combined norm that is in the form of weighted $\ell_1$ norm. We will see later that this facilitates the updating of coefficients matrix $C$ when solving the optimization problem.

\item {\mcb Many theoretical results have been established for SSC, which show that it is able to find subspace-preserving solution when subspaces are independent~\cite{Elhamifar:CVPR09,Elhamifar:TPAMI13}, disjoint~\cite{Elhamifar:ICASSP10}, and affine~\cite{Li:SPARS15}, and when data are corrupted by outliers~\cite{Soltanolkotabi:AS12}, contaminated by noise~\cite{Wang-Xu:ICML13,Soltanolkotabi:AS14,Wang:JMLR16}, and preprocessed using dimension reduction~\cite{Wang:ICML15}.}

\end{itemize}

Equipped with the \textit{subspace structured} $\ell_1$ norm of $C$, we can reformulate the unified optimization framework in~\eqref{eq:STSC-outliers} for subspace clustering as follows:
\begin{align}
\begin{split}
\min\limits_{C,E,Q} ~&
\|C\|_{1,Q} + \lambda \|E\|_{\E} \\
\st ~~ &X = X C + E, ~\text{diag}(C) = \0, ~~ Q \in \Q,
\end{split}
\label{eq:SSSC-outliers}
\end{align}
where the norm $\|\cdot\|_\E$ on the error term $E$ depends upon the prior knowledge about the pattern of noise or corruptions.\footnote{The $\ell_{2,1}$, Frobenius, $\ell_1$ norms are used for gross corruptions over a few columns, dense Gaussian noise, and sparse corruptions, respectively. A combination of them are used for mixed patterns of noise and corruptions.}
We call problem \eqref{eq:SSSC-outliers} as \textit{Structured Sparse Subspace Clustering} (SSSC or \s3c).

\myparagraph{Remark 1} The \s3c framework in \eqref{eq:SSSC-outliers} generalizes SSC because, instead of first solving for a sparse representation to find $C$ and then applying spectral clustering to the affinity $|C|+|C^\top|$ to obtain the segmentation, in \eqref{eq:SSSC-outliers} we simultaneously search for the sparse representation $C$ and the segmentation $Q$. To be more specific, when the parameter $\alpha$ in $\|C\|_{1,Q}$ is set as $0$ or $Q$ is initialized as a noninformative matrix, \eg,  all zeros, 
problem \eqref{eq:SSSC-outliers} for $(C, E)$ reduces to SSC. Thus, we find an initialization for $(C,E)$ by solving 
problem \eqref{eq:SSSC-outliers} with $Q=\0$ and $\alpha > 0$ which reduces to SSC.


\myparagraph{Remark 2} Our \s3c differs from the reweighted $\ell_1$ minimization \cite{Candes:JFAA08} in that what we use to reweight is not $C$ itself but rather a segmentation matrix $Q$. The attempt to integrate these two stages into a unified framework is also suggested in the work of Feng \etal~\cite{Feng:CVPR14}, who introduce a block-diagonal constraint into the self-expressiveness model. However, this requires precise knowledge of the segmentation, and therefore enforcing exact block-diagonality is not possible with their model.
Compared to \cite{Feng:CVPR14}, which adds an extra block-diagonal constraint into the expressiveness model, 
our framework encourages consistency between the representation matrix and the estimated segmentation matrix. It should also be noted that our subspace structured norm can be used in conjunction with other self-expressiveness based methods \cite{Liu:ICML10,Lu:ECCV12,Lu:ICCV13-TraceLasso,Lu:ICCV13,Guo:IJCAI15}, with other weighted methods \cite{Pham:CVPR12, Hu:CVPR14} or dictionary learning \cite{Peng:CVPR13, Kodirov:ECCV16}, and can also be extended to deal with missing entries \cite{Li:TSP16}.

\subsection{Constrained Structured Sparse Subspace Clustering: An Extension to Incorporate Side Information}
\label{sec:incorp-prior-info}

In some applications, for example, in the task of clustering genes in DNA microarray data, there often exists prior 
knowledge about the relationships between some subset of genes or genes expression profiles \cite{Fang:JBI06, Chopra:BMCbioinfo08, Huang:Bioinfo06, Bair:WIRCS13}. This prior knowledge essentially provides partial side-information to indicate ``must-link'' or ``cannot-link'' constraints in clustering.

The direct way to incorporate side-information into our \s3c framework is to partially initialize the structure matrix $\Theta$ with the ``cannot-link'' constraints in the side-information. Here, we propose to incorporate the side-information to weight the $\ell_1$ norm, that is, to modify $\| C \|_1 =\|C \odot \1 \1^\top \|_1$ into $\| C \|_{\Psi} = \| C \odot \Psi \|_1$, where the operator $\odot$ is the Hadamard product (\ie, element-wise product) and $\Psi$ encodes the side-information as follows:
\begin{itemize}
\item $\Psi_{ij} = \exp{(-1)}$ if data points $i$ and $j$ have a ``must-link'', \ie, data points $i$ and $j$ should belong to the same subspace;
\item $\Psi_{ij} = \exp{(+1)}$ if data points $i$ and $j$ have a ``cannot-link'', \ie, data points $i$ and $j$ should belong to different subspaces;
\item $\Psi_{ij} = 1$ otherwise, \ie, there is no side-information for data points $i$ and $j$.
\end{itemize}

Consequently, the subspace structured $\ell_1$ norm $\|C\|_{1,Q}$ can be modified to incorporate the side-information $\Psi$ as follows:
\begin{align}
\label{eq:subspace-structured-L-1-Q-Psi}
\begin{split}
\|C\|_{\Psi,Q} &\doteq  \| C \odot \Psi\|_1 + \alpha \|C\|_Q \\
&= \sum_{i,j}|C_{ij}|(\Psi_{ij} + \frac{\alpha}{2}\|\q^{(i)} - \q^{(j)}\|^2).
\end{split}
\end{align}
Note that if the side-information is not available, $\|C\|_{\Psi,Q}$ reduces to $\|C\|_{1,Q}$ because the side-information matrix $\Psi$ degenerates to $\1 \1^\top$.
Therefore, we extend the structured sparse subspace clustering problem to incorporate the given side-information by solving the following problem:
\begin{align}
\begin{split}
\min\limits_{C,E,Q} ~&
\|C\|_{\Psi,Q} + \lambda \|E\|_{\E} \\
\st ~~ &X = X C + E, ~\text{diag}(C) = \0, ~~ Q \in \Q.
\end{split}
\label{eq:SSSC-outliers-side-info}
\end{align}

We call problem \eqref{eq:SSSC-outliers-side-info} as \textit{Constrained Structured Sparse Subspace Clustering} (CSSSC or \cs3c).

\myparagraph{Remark 3} While there are more sophisticated strategies to incorporate the side-information, the reason to adopt a simple weighting approach as in problem \eqref{eq:SSSC-outliers-side-info} is that by doing so we can solve the problem with minor changes 
to the optimization algorithm, as we will show at the end of Section \ref{sec:spectral-clustering}.

\section{Alternating Minimization Algorithms for Structured Sparse Subspace Clustering}
\label{sec:solving-algorithm}

In this section, we present algorithms to solve the optimization problem \eqref{eq:SSSC-outliers} by alternating between the following two subproblems:
\begin{enumerate}
\item Find $C$ and $E$ given $Q$ by solving a subspace structured sparse representation problem.
\item Find $Q$ given $C$ and $E$ by spectral clustering. 
\end{enumerate}

One way to solve problem \eqref{eq:SSSC-outliers} as explored in \cite{Li:CVPR15} is to alternate between computing the subspace structured sparse representation and applying spectral clustering. Specifically, in spectral clustering the discrete constraint $Q \in \Q$ is relaxed so that the optimal $Q$ can be computed from the singular value decomposition of the graph Laplacian, then $Q$ is quantized 
into the valid segmentation set $\Q$ by applying the $k$-means algorithm. We call this procedure as \textit{hard} \s3c.

In hard \s3c, a binary segmentation matrix $Q$ is used to re-weight the updating of $C$ in the next iteration. While using a binary segmentation matrix $Q$ is conceptually simple, it may not be capable of capturing the detailed information in the clustering results. Notice that, some data points are easier to segment and thus their clustering results are more reliable, whereas some data points are harder to segment and thus their clustering results are less reliable. Such kind of detailed confidence or uncertainty information 
would be ignored when quantizing the clustering results into a binary segmentation matrix $Q$.
%
%
Therefore, instead of quantizing 
$Q$ by using $k$-means in each iteration, we propose to use the real-valued matrix $Q \in \RR^{N\times n}$ for re-weighting the updating of $C$ 
in the next iteration. We call this procedure as \textit{soft} \s3c. When compared to using a binary $Q$ in hard \s3c, the continuous real-valued $Q$ in soft \s3c carries more detailed information of the previous clustering results and thus re-weights the updating of representation matrix $C$ smoothly. This is beneficial to yield better clustering results.

The optimization problem in \eqref{eq:SSSC-outliers-side-info} can be solved with some minor changes in the algorithms for solving problem \eqref{eq:SSSC-outliers}.

\subsection{Subspace Structured Sparse Representation}
\label{sec:subspace-structured-SR}

Given a binary segmentation matrix $Q$ or real-valued matrix $Q$, we compute the subspace structure matrix $\Theta$ where $\Theta_{ij}\doteq\frac{1}{2}\|\q^{(i)} - \q^{(j)}\|_2^2$ in which $\q^{(i)}$ and $\q^{(j)}$ are the $i$-th and $j$-th rows of matrix $Q$, respectively. Then we solve for $C$ and $E$ by solving the following subspace structured sparse representation problem
\begin{align}
\label{eq:ST-SSR}
\!\!
\begin{split}
\min\limits_{C,E} ~& \|C\|_{1,Q} + \lambda \|E\|_{\E}  \\
\st ~~ &X = X C + E, ~ \diag(C) = \0, \!\!
\end{split}
\end{align}
which is equivalent to the following problem
\begin{align}
\label{eq:ST-SSR-C}
\min\limits_{A,C,E} ~&  \|C\|_{1,Q} + \lambda \|E\|_{\E} \\
\st ~~ & X = X A + E,  A = C - \diag(C).\nonumber
\end{align}
We solve this problem using the Alternating Direction Method of Multipliers (ADMM) \cite{Lin:09}, \cite{Boyd:FTML10}. 
The augmented Lagrangian is given by:
\begin{align}
\begin{split}
\!\!\!\!\mathcal{L}&(C,A,E,Y,Z) \\
=&\|C\|_{1,Q} + \lambda \|E\|_{\E} + \langle Y, X-XA-E \rangle \\
+&\langle Z, A - C + \text{diag}(C)\rangle \\
+&\frac{\mu}{2}(\|X - XA - E\|^2_F + \|A - C + \text{diag}(C)\|^2_F),
\end{split}
\label{eq:ST-SSR-ALM}
\end{align}
where $Y$ and $Z$ are matrices of Lagrange multipliers, and $\mu > 0$ is a parameter. To find a saddle point for $\mathcal{L}$, we update each of $C$, $A$, $E$, $Y$, and $Z$ alternately while keeping the other variables are fixed.

\begin{algorithm}[tb]
   \caption{\bf (ADMM for solving problem \eqref{eq:ST-SSR-C})}
   \label{alg:ST-SSR-ADMM}
\begin{algorithmic}
      \REQUIRE Data matrix $X$, and parameters $\lambda$ and $\alpha$. 
      \STATE {\bfseries Initialize:} $\Theta=\0$, 
      $E^{(0)} = \0$, $A^{(0)}=\0$, 
      $Y^{(0)}=\0$, $Z^{(0)}=\0$, $\epsilon=10^{-6}$, $\rho=1.1$, $\mu^{(0)}=\frac{1}{\min_j \max_{i:i \neq j} \{\x_i^\top \x_j\}}$, $t=0$.
      \label{alg:init}
    \WHILE{not converged}
        \STATE Update $C^{(t+1)}$, $A^{(t+1)}$, and $E^{(t+1)}$;
        \STATE Update $Y^{(t+1)}$ and $Z^{(t+1)}$;
        \STATE Update $\mu^{(t+1)} \leftarrow \rho \mu^{(t)}$;
        \STATE Check the convergence condition $\|X-XA^{(t+1)}-E^{(t+1)}\|_\infty < \epsilon$; if not converged, then set $t\leftarrow t+1$.
    \ENDWHILE
    \ENSURE $C^{(t+1)}$ and $E^{(t+1)}$
\end{algorithmic}
\end{algorithm}
%

\myparagraph{Update for $C$} We update $C$ by solving the following problem:
\begin{align}
\!\!\!\!C^{(t+1)}\! = \!\argmin \limits_{C} & \frac{1}{\mu^{(t)}} \|C\|_{1, Q} + \frac{1}{2} \|C - \text{diag}(C) - U^{(t)}\|^2_F,\!\! \nonumber
\end{align}
where $\|C\|_{1, Q}= \|C\|_1 + \alpha\|\Theta \odot C\|_1$ and $U^{(t)} = A^{(t)} + \frac{1}{\mu^{(t)}}Z^{(t)}$. The closed-form solution for $C$ is given as
\begin{align}
C^{(t+1)} = \tilde C^{(t+1)} - \diag(\tilde C^{(t+1)}),
\label{eq:eliminate-C-diagonal}
\end{align}
where the $(i,j)$ entry of $\tilde C$ is given by
\begin{align}
\tilde C^{(t+1)}_{ij} = \S_{\frac{1}{\mu^{(t)}}(1 + \alpha\Theta_{ij})} (U^{(t)}_{ij}),
\label{eq_SLR_SMC_Z_solu}
\end{align}
where $\S_\tau(\cdot)$ is the shrinkage thresholding operator. 
Note that the subspace structured $\ell_1$ norm causes a minor change to the algorithm used to compute $C$ from the standard SSC --- because of the homogeneity in the two terms. Namely, rather than soft-thresholding all the entries of matrix $U^{(t)}$ uniformly with a constant value, we threshold each entry $U^{(t)}_{ij}$ of matrix $U^{(t)}$ with a different value that depends on $\Theta_{ij}$.

\myparagraph{Update for $A$}
We update $A$ by solving 
problem:\footnote{Note that $\text{diag}(C^{(t+1)})$ is dropped because it is $\0$ during iterations due to the updating rule in \eqref{eq:eliminate-C-diagonal}.}
\begin{align}
\begin{split}
\!\!\!\!A^{(t+1)}& = \argmin \limits_{A} \langle Y^{(t)}, X-XA-E^{(t)} \rangle \\
     + & \langle Z^{(t)}, A -C^{(t+1)}\rangle \\ 
     + & \frac{\mu^{(t)}}{2}(\|X\!-\!XA\!-\!E^{(t)}\|^2_F + \|A\!-\!C^{(t+1)}\|^2_F),\nonumber 
\end{split}
\end{align}
whose solution is given by
\begin{align}
\begin{split}
A^{(t+1)}
\!&=\!(X^\top X \!+\! \I)^{-1}[X^\top (X \!-\! E^{(t)} \!-\! \frac{1}{\mu^{(t)}}Y^{(t)})
\! \\
  & +\! C^{(t+1)} \!-\! \frac{1}{\mu^{(t)}}Z^{(t)}]. 
\end{split}
\label{eq_SLR_SMC_C_solu}
\end{align}

\myparagraph{Update for $E$} While other variables are fixed, we update $E$ as follows:
\begin{equation}
E^{(t+1)} =\argmin \limits_{E} {\hspace{-0pt}} \frac{\lambda}{\mu^{(t)}} \|E\|_\E +\frac{1}{2}\|E - V^{(t)}\|^2_F
\label{eq_SLR_SMC_Es}
\end{equation}
where $V^{(t)}= X-XA^{(t+1)} +\frac{1}{\mu^{(t)}}Y^{(t)}$. If we use the $\ell_1$ norm for $E$, then
\begin{equation}
E^{(t+1)} =\S_{\lambda \over \mu^{(t)}} ( V^{(t)}).
\label{eq_SLR_SMC_E}
\end{equation}

\myparagraph{Update for $Y$ and $Z$} The update for the Lagrange multipliers is a simple gradient ascent step
\begin{align}
\begin{split}
Y^{(t+1)} &=Y^{(t)} + \mu^{(t)}(X - XA^{(t+1)}-E^{(t+1)}), \\
Z^{(t+1)} &=Z^{(t)} + \mu^{(t)}(A^{(t+1)} - C^{(t+1)}).\nonumber 
\end{split}
\end{align}

For clarity, we summarize the ADMM algorithm for solving problem~\eqref{eq:ST-SSR-C} in Algorithm \ref{alg:ST-SSR-ADMM}. For the details of the derivation, we refer the readers to \cite{Lin:09, Boyd:FTML10}.

\subsection{Spectral Clustering}
\label{sec:spectral-clustering}

Given $C$ and $E$, problem \eqref{eq:SSSC-outliers} reduces to the following problem:
\begin{eqnarray}
\begin{array}{rl}
\min\limits_{Q} \| C \|_{Q} ~~\text{ s.t.} ~~ Q \in \Q. 
\end{array}
\label{eq:solving-Q-from-Z-1-Q}
\end{eqnarray}

By using the definition of the subspace structured norm in \eqref{eq:structured-Q-norm}, problem \eqref{eq:solving-Q-from-Z-1-Q} can be identified as the spectral clustering problem
\begin{eqnarray}
\begin{array}{rl}
\min\limits_{Q} \trace(Q^\top \bar L Q) ~~\text{ s.t.} ~~ Q \in \Q,
\end{array}
\end{eqnarray}
in which $\bar L$ is the graph Laplacian of the data affinity matrix $\bar A$ 
defined in \eqref{eq:def-Affinity}, \ie, $\bar L = \bar D - \bar A$, $\bar A = \frac{1}{2}(|C|+|C^\top|)$, and $\bar D$ is the degree matrix with diagonal entries $\bar D_{jj} = \sum_i \bar A_{ij}$.
In particular, when relaxing the constraint $Q \in \Q$ to the constraint $Q^\top \bar D Q =I$,
we obtain:
\begin{align}
\min_{Q \in \RR^{N\times n}}  ~\trace(Q^\top \bar L Q) \quad \st \quad Q^\top \bar D Q = I,
\label{eq:SC-Lap-QtDQ}
\end{align}
then the solution $Q \in \Q$ can be found by spectral clustering with normalized cut \cite{Shi-Malik:PAMI00}. More specifically, let $\tilde Q = \bar D^{\frac{1}{2}} Q $, then problem \eqref{eq:SC-Lap-QtDQ} turns out to be:
\begin{align}
\min_{\tilde Q \in \RR^{N\times n}}  ~\trace(\tilde Q^\top \bar D^{-\frac{1}{2}}\bar L \bar D^{-\frac{1}{2}} \tilde Q) \quad \st \quad \tilde Q^\top \tilde Q = I.
\label{eq:SC-normalized-Lap-tildeQ}
\end{align}
The solution of $\tilde Q$ is given by the bottom $n$ eigenvectors of $\bar D^{-\frac{1}{2}}\bar L \bar D^{-\frac{1}{2}}$ associated with its $n$ smallest eigenvalues.

\begin{itemize}
\item In hard \s3c, the rows of $\tilde Q \in \RR^{N \times n}$ are used as the input to the $k$-means algorithm and the clustering of the rows of $\tilde Q$ is used to define the solution of $Q \in \Q$ as $q_{ij} = 1$ if point $i$ belongs to cluster $j$, and $q_{ij} = 0$ otherwise.
    Then, we construct a binary subspace structure matrix $\Theta$, where $\Theta_{ij}=\frac{1}{2}\|\q^{(i)} - \q^{(j)}\|^2_2 \in \{0, 1\}$.

\item In soft \s3c, we use the matrix $\tilde Q \in \RR^{N \times n}$ to construct the subspace structure matrix $\Theta$, where each row of $\tilde Q$ is normalized to unit $\ell_2$ norm and $\Theta_{ij}=\frac{1}{2}\|\q^{(i)} - \q^{(j)}\|^2_2 \in [0, 2]$.
\end{itemize}

\subsection{Algorithm Summary}
The \s3c algorithms alternate between solving for the matrices of sparse coefficients and error $(C, E)$ 
given the segmentation $Q$ using Algorithm~\ref{alg:ST-SSR-ADMM} and solving for $Q$ given $(C,E)$ using spectral clustering. The hard \s3c defines a binary subspace structure matrix $\Theta$ with the clustering indicator matrix $Q$; whereas the soft \s3c defines a real-valued subspace structure matrix $\Theta$ with matrix $Q \in \RR^{N\times n}$. For clarity, we summarize the procedure for solving problem~\eqref{eq:SSSC-outliers} in Algorithm~\ref{alg:hard-S3C}. {\mcb As the affinity matrix is sparse, the main computational burden of Algorithm~\ref{alg:hard-S3C} is in solving problem \eqref{eq:ST-SSR}. Specifically, the computational cost is $O(T_1T_2(N^3+DN^2))$ due to the matrix inverse and matrix multiplication in updating $A$ by \eqref{eq_SLR_SMC_C_solu}, where $T_1$ is the number of iteration in solving \eqref{eq:ST-SSR} with ADMM and $T_2$ is the number of outer iterations in Algorithm~\ref{alg:hard-S3C}. Note that, in Algorithm \ref{alg:hard-S3C}, the previous solution is used to initialize the next execution of ADMM and thus starting from the second iteration in Algorithm \ref{alg:hard-S3C}, $T_1$ could be remarkably reduced.
}

\myparagraph{Remark 4}
To solve \cs3c in \eqref{eq:SSSC-outliers-side-info}, we change the step for updating $\tilde C^{(t+1)}_{ij}$ in \eqref{eq_SLR_SMC_Z_solu} to the following:
\begin{align}
\tilde C^{(t+1)}_{ij} = \S_{\frac{1}{\mu^{(t)}}(\Psi_{ij} + \alpha\Theta_{ij})} (U^{(t)}_{ij}).
\label{eq_SLR_SMC_Z_solu_side_info_Psi}
\end{align}
Note that the side-information encoded in $\Psi$ imposes 
``supervision'' on some entries of $\tilde C^{(t)}$: 
a) the entry $\tilde C^{(t)}_{ij}$ is penalized more heavily if having a ``cannot-link'' constraint, and b) the entry $\tilde C^{(t)}_{ij}$ is encouraged if having a ``must-link'' constraint.

\begin{algorithm}[!tb]
   \caption{(\s3c)} 
   \label{alg:hard-S3C}
\begin{algorithmic}
   \REQUIRE Data $X$, number of subspaces $n$, parameters $\lambda, \alpha$
   \STATE {\bfseries Initialize} $(C, E)$ by SSC
   \WHILE {not converged} 
   \STATE Given $(C,E)$, solve problem \eqref{eq:solving-Q-from-Z-1-Q} via spectral clustering to obtain $Q$;
   \STATE Given $Q$, solve problem \eqref{eq:ST-SSR} via Alg.~\ref{alg:ST-SSR-ADMM} to obtain $(C,E)$;
   \ENDWHILE
   \ENSURE Segmentation matrix $Q$
\end{algorithmic}
\end{algorithm}


\section{Discussion on Effects of Using $\|C\|_{Q}$, Role of Parameter $\alpha$, and Stopping Criterion}
\label{sec:discussions}

\subsection{Effects of Using Subspace Structured Norm}
\label{sec:effect-of-subspace-structured-L1-norm}

{\mcb By using the subspace structured norm $\|C\|_{Q}$, the interaction between the representation matrix $C$ and the segmentation matrix $Q$ is captured. More specifically, the alternating minimization strategy in Algorithm \ref{alg:hard-S3C} performs effectively a multi-round refinement procedure for learning a subspace-preserving solution $C$ and a correct segmentation matrix $Q$.}
\begin{itemize}
\item Suppose that the optimal solution $C$ is subspace-preserving, then the subspace structured norm $\|C\|_{Q}$ vanishes whenever $Q$ is the correct segmentation matrix\footnote{Our discussion here assumes that $Q$ is binary.}.

\item Suppose that the optimal solution $C$ does not satisfy the subspace-preserving property 
exactly but spectral clustering still yields a correct subspace clustering, then $C$ would be refined towards being subspace-preserving in the next iteration.

\item Suppose that the optimal solution $C$ does not satisfy the subspace-preserving property 
exactly and spectral clustering does not yield a correct subspace clustering. In this case, the subspace segmentation matrix obtained in the previous step will refine the representation matrix and is also able to facilitate a better initialization for the next running of clustering.

\end{itemize}

In the first two cases, the optimal solution $C_\ast$ obtained by \s3c would not be worse than that from the original SSC. In the third case,
we have no guarantee yet that the optimal solution $C_\ast$ obtained by \s3c must improve over the original SSC.
In experiments, we observe that the clustering accuracy, the connectivity in the obtained affinity matrix, and the subspace-preserving property of the coefficient matrix $C$ of \s3c improve over the original SSC when the parameters are set properly.

\subsection{Role of Parameter $\alpha$ in \s3c}
\label{sec:role-of-parameter-alpha}

Recall that the subspace structured $\ell_1$ norm $\|C\|_{1,Q}$ uses a tradeoff parameter $\alpha$, \ie, $ \|C\|_{1,Q} = \| C \|_1 + \alpha \|C\|_Q$. During iterations of \s3c, the parameter $\alpha$ balances how much we trust the previous segmentation matrix $Q$ and the current SSC. To be more specific, a small $\alpha$ will give a solution of $C$ that is close to that of SSC, whereas a large $\alpha$ will give a solution for $C$ that is more dependent on the previous segmentation matrix $Q$. By using an 
appropriate $\alpha$, we can allow SSC to provide a good initialization and then let the iterations improve based on better estimates of the segmentation matrix $Q$. In experiments, we will evaluate the clustering performance of \s3c with respect to $\alpha$ and the effects of using the subspace structured norm $\|C\|_Q$ during iterations.

We observed that rather than using a fixed $\alpha$ to balance the two terms in the subspace structured $\ell_1$ norm $\|C\|_{1,Q}$, the performance 
could be 
improved by, \eg, using $\alpha \leftarrow \nu \alpha$,
or using $\nu^{1-T} \|C\|_1 + \alpha \nu^{T-1} \|C \|_Q$ in Algorithm~\ref{alg:hard-S3C}, where $\nu >1$ and $T=1,2,\dots$ is the iteration index.
By doing so, the previously estimated segmentation matrix $Q$ (or subspace structure $\Theta$) is increasingly emphasized during the iterations. In experiments, we will evaluate the performance of \s3c with respect to three strategies to set the parameter $\alpha$ as follows:
\begin{itemize}
\item Using a fixed $\alpha$, \ie, using $\|C\|_{1,Q} = \|C \|_1 + \alpha \|C \|_Q$.
\item Increasing $\alpha$ gradually during iterations, \ie, using $\|C\|_{1,Q} = \|C \|_1 + \alpha \nu^{T-1} \|C \|_Q$, where $\nu >1$. 
\item Increasing $\alpha$ and at the same time decreasing the $\ell_1$ norm term gradually during iterations, \ie, using $\nu^{1-T} \|C \|_1 + \alpha \nu^{T-1} \|C \|_Q$, where $\nu > 1$. 
\end{itemize}

\subsection{Stopping Criterion}
While the problem solved by Algorithm~\ref{alg:ST-SSR-ADMM} is a convex problem, there is no guarantee that the Algorithm~\ref{alg:hard-S3C} will converge to a global or local optimum because the solution for $Q$ given $(C,E)$ is obtained in an approximate manner by relaxing the objective function. Nonetheless, our experiments show that the algorithm does converge for proper settings of the parameters.

In practice, Algorithm~\ref{alg:hard-S3C} can be stopped by setting a maximum iteration number $T_{\text{max}}$, or when the relative changes of subspace structure matrix $\Theta$ or coefficients matrix $C_\ast$ in two consecutive iterations is small enough, \ie,
\begin{align}
\label{eq:Alg2-converge-stopping-rules-I}
&\frac{\|\Theta^{(T)} - \Theta^{(T+1)}\|_1}{\|\Theta^{(T)}\|_1} < \epsilon_1, \\
\label{eq:Alg2-converge-stopping-rules-II}
&\frac{\|C_\ast^{(T)} - C_\ast^{(T+1)}\|_1}{\|C_\ast^{(T)}\|_1} < \epsilon_2,
\end{align}
where $C_\ast^{(T)}$ is the optimal solution of $C$ in problem \eqref{eq:solving-Q-from-Z-1-Q} and $\Theta^{(T)}$ is the structure matrix $\Theta$ at the $T$-th iteration of Algorithm~\ref{alg:hard-S3C}, respectively, in which $T=1,2,\dots$, $\epsilon_1 >0$ and $\epsilon_2 >0$.

In addition\footnote{Besides, the decrease of the optimal cost of $k$-means in spectral clustering between two consecutive iterations in Algorithm~\ref{alg:hard-S3C} can also be used as a stopping rule, \ie, we can stop Algorithm~\ref{alg:hard-S3C} if
\begin{align}
\delta_\ast^{(T)} - \delta_\ast^{(T+1)} < \epsilon_4,
\label{eq:Alg2-converge-stopping-rules-IV}
\end{align}
where $\delta_\ast^{(T)}$ is the optimal cost of $k$-means at iteration $T$ of Algorithm~\ref{alg:hard-S3C} and $\epsilon_4 >0$.}, the subspace structured norm of $C$ with respect to $Q$, \ie, $\|C\|_Q$, is assumed to be monotonously decreasing. Thus, we can also stop Algorithm~\ref{alg:hard-S3C} when the decrease of $\|C_\ast\|_Q$ between two consecutive iterations is small enough, \ie,
\begin{align}
&\|C_\ast^{(T)}\|_Q - \|C_\ast^{(T+1)}\|_Q < \epsilon_3, 
\label{eq:Alg2-converge-stopping-rules-III}
\end{align}
where $\epsilon_3 >0$.

\myparagraph{Remark 5} When a fixed $\alpha$ is used, the relative changes in the objective function of problem \eqref{eq:SSSC-outliers}, the relative changes in subspace structure matrix $\Theta$ as defined in \eqref{eq:Alg2-converge-stopping-rules-I}, and the relative changes in coefficients matrix $C_\ast$ as defined in \eqref{eq:Alg2-converge-stopping-rules-II} are useful to stop Algorithm \ref{alg:hard-S3C}. If parameter $\alpha$ is changed during iterations, however, the relative changes in objective function and the relative changes in coefficients matrix $C_\ast$ would no longer be valid to stop Algorithm \ref{alg:hard-S3C}. 
As a remedy, we can use the stopping rule in \eqref{eq:Alg2-converge-stopping-rules-IV} or \eqref{eq:Alg2-converge-stopping-rules-III}, which virtually performs \textit{model selection} for subspace clustering.

\begin{table*}[tb]
\caption{Clustering Errors on Synthetic Data Set. The best results are in bold font.} 
\centering
\small
\begin{tabular}{|c|c|c|c|c|c|c|c|c|c|c|}
\hline
Corruptions (\%)      & 0    & 10\%     & 20\%      &  30\%     & 40\%     &50 \%    & 60\%   & 70\%    & 80 \%   & 90\% \\
\hline
\hline
SSC       & 1.43 &1.93 & 2.17 &4.27 &16.87 &32.50 &54.47 &62.43 & 68.87 & 73.77        \\
\textbf{hard \s3c} & \textbf{0.30} & \textbf{0.33} & \textbf{0.90} & \text{2.97} &\text{10.70}     & \textbf{23.67}    &\textbf{50.50}   & \textbf{60.70} & \textbf{67.97} & \textbf{73.33} \\
\textbf{soft \s3c} & \text{0.73} & \text{0.90} & \text{1.20} & \textbf{2.60} &\textbf{10.40}     & \text{24.90}    &\text{50.57}   & \text{61.93} & \text{70.10} & \text{73.70} \\
\hline
\end{tabular}
\label{table:synthetic-data-SSC-vs-SSSC}
\end{table*}

\section{Experiments}
\label{sec:experiments}
In this section, we evaluate the effectiveness of the proposed \s3c approach on a synthetic data set, a face clustering data set, a motion segmentation data set, and three cancer gene expression data sets.

\myparagraph{Experimental Setup} Since that \s3c is 
a generalization of the standard SSC \cite{Elhamifar:TPAMI13}, we keep all settings in \s3c the same as that in SSC and thus the first iteration of hard \s3c is equivalent to a standard SSC. The \s3c specific parameters are set using $\nu^{1-T} \|C\|_1 + \alpha \nu^{T-1} \|C \|_Q$, where $\nu=1.2$ and $T_{max}=10$ by default. The ADMM parameters are kept the same for both \s3c and SSC.
{\mcb For each algorithm, we assume that the number of clusters $n$ is known. The parameter $\lambda$ in SSC is set by $\lambda=\frac{\lambda_0}{\min_j \max_{i:i \neq j} \{\x_i^\top \x_j\}}$ where $\lambda_0$ is tuned on each data set. In \s3c, we keep $\lambda$ the same as in SSC and tune $\alpha$ on each data set. By default, we set $\alpha=0.1$ in hard \s3c, and $\alpha=1$ in soft \s3c.}
For each set of experiments, the mean and median (or standard deviation) of subspace clustering error are recorded. The error (ERR) of subspace clustering is calculated by
\begin{equation}
ERR(\a, \hat \a) = 1- \max \limits_\pi \frac{1}{N}\sum_{i=1}^N 1_{\{\pi(a_i) = \hat a_i\}}
\end{equation}
where $\a, \hat \a \in \{1,\cdots, n\}^N$ are the original and estimated assignments of the columns in $X$ to the $n$ subspaces, and the maximum is with respect to all permutations
\begin{align}
\pi:\{1,\cdots, n\}^N\rightarrow \{1,\cdots, n\}^N.
\end{align}

\begin{figure*}[tb!]
\centering
\subfigure[ERR]{\includegraphics[clip=true,trim=0 0 0 0,width=0.575\columnwidth]{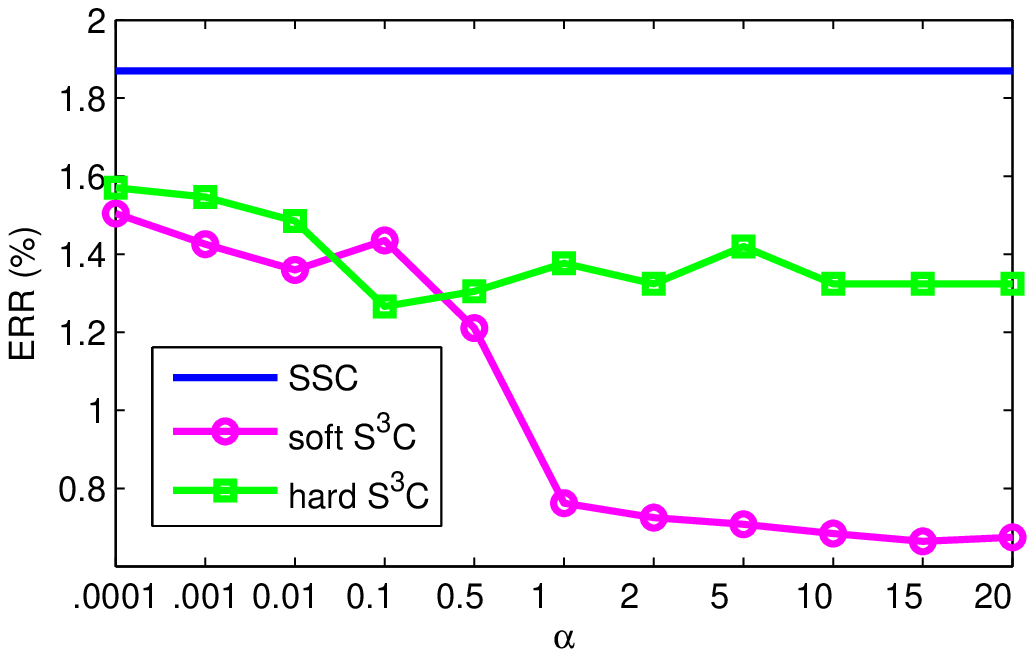}} 
\subfigure[CONN]{\includegraphics[clip=true,trim=0 0 0 0,width=0.575\columnwidth]{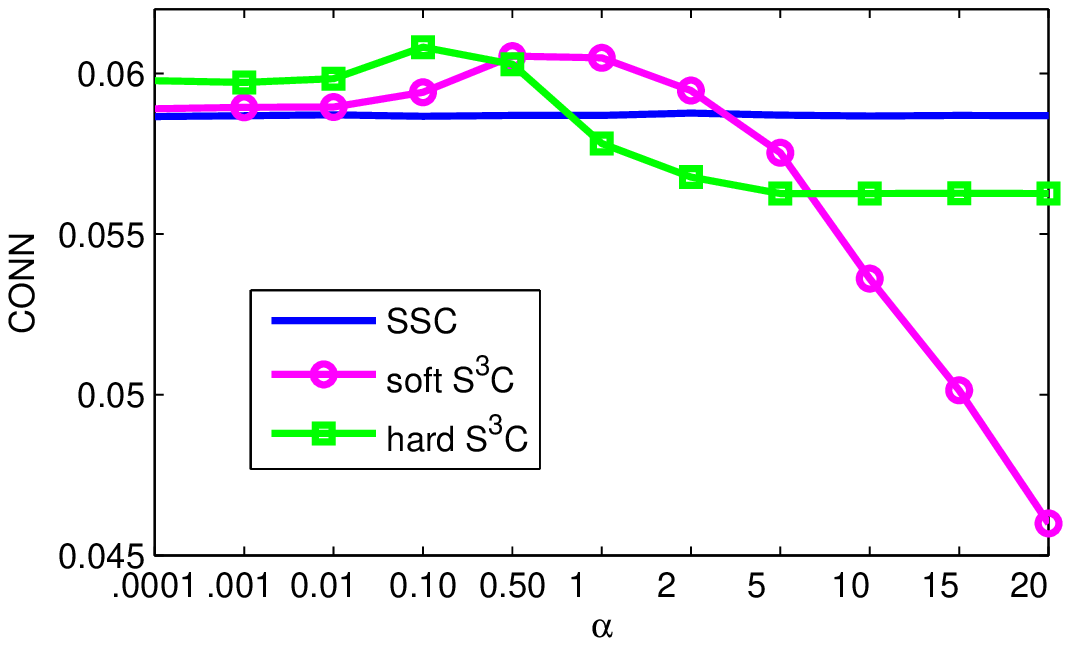}}
\subfigure[SPR]{\includegraphics[clip=true,trim=0 0 0 0,width=0.575\columnwidth]{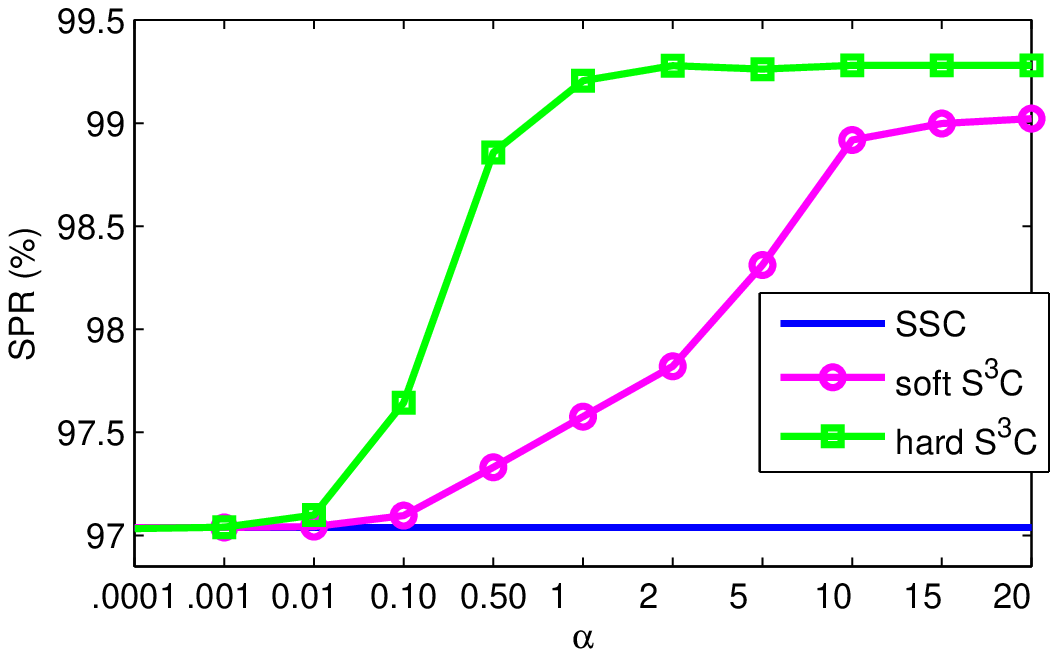}}
\caption{Experimental evaluation on the behavior of \s3c with varying value of parameter $\alpha$. Panels (a)-(c) show the clustering error (ERR), the connectivity (CONN) and the subspace-preserving rate (SPR) of the affinity matrix as a function of parameter $\alpha$, respectively.}
%
%
\label{fig:ERR-CONN-SRR-ZQnorm-vs-alpha}
\end{figure*}

In addition to ERR, we also report the subspace-preserving rate (SPR) and the graph connectivity (CONN), which are defined as follows:
\begin{itemize}
\item SPR quantifies the degree to which the solution $C$ satisfies the subspace-preserving property. For each column $\c_j$ of $C$, we compute the fraction of its $\ell_1$ norm that comes from the correct subspaces and average over all $j$, \ie,
    \begin{equation}
    \text{SPR} = \frac{1}{N}\sum_j (\sum_i(w_{ij}\cdot |c_{ij}|) / \|\c_j\|_1),
    \end{equation}
    where $w_{ij} \in \{0, 1\}$ is the ground truth affinity. By this definition, $\text{SPR} \in [0, 1]$ and is $1$ if and only if $C$ is subspace-preserving.

\item CONN 
    evaluates the connectivity of the affinity graph generated from the proposed method.
    Generally, for an undirected graph with affinity matrix $A\in \RR^{N\times N}$, the graph connectivity is defined as the second smallest eigenvalue of the Laplacian $L = \I - D^{-1/2} A D^{-1/2}$ where $D = \diag(A \cdot \1)$. The connectivity is in the range of $[0, \frac{N-1}{N}]$ and is zero if and only if the graph is not connected \cite{Fiedler:CMathJ1975}. 
    We define CONN of an affinity matrix $A$ as the 
    average of CONN of the sub-graph corresponding to the points in one of the subspaces.

\end{itemize}

\subsection{Experiments on Synthetic Data}
\label{sec:experiments-synthetic-data}

\myparagraph{Data Preprocessing}
We construct $n$ linear subspaces $\{S_j\}^n_{j=1}\subset \RR^{D}$ of dimension $d$ by choosing their bases $\{U_j\}^n_{j=1}$ 
as the top $d$ left singular vectors of a random matrix $R_j \in \RR^{D\times D}$. We sample $N_j$ data points from each subspace $j=1,\dots,n$ as $X_j = U_j Y_j$, where the entries of $Y_j\in\RR^{d\times N_j}$ are i.i.d. samples from a standard Gaussian. Note that by doing so, the $n$ linear subspaces are not necessarily orthogonal to each other. We then corrupt a certain percentage $p = 10-90\%$ of the entries of $X_j$ chosen uniformly at random by adding Gaussian noise with zero mean and variance $0.3\|\x\|$ to the selected entries. In our experiments, we set $D=100$, $d=5$, $N_j=10$, $n=15$, and repeat each experiment for $20$ trials.

Experimental results are presented in Table~\ref{table:synthetic-data-SSC-vs-SSSC}. We observe that both the hard \s3c and the soft \s3c algorithms consistently outperform SSC. The improvements of \s3c from SSC are relatively more significant for the corruption in the range $10-50\%$. Compared to soft \s3c, hard \s3c performs slightly better. These results confirm the robustness and effectiveness of the proposed joint optimization approach.

\begin{figure*}[tb!]
\centering
\subfigure[$C^{(1)}$]{\includegraphics[clip=true,trim=20 10 30 15,width=0.50\columnwidth]{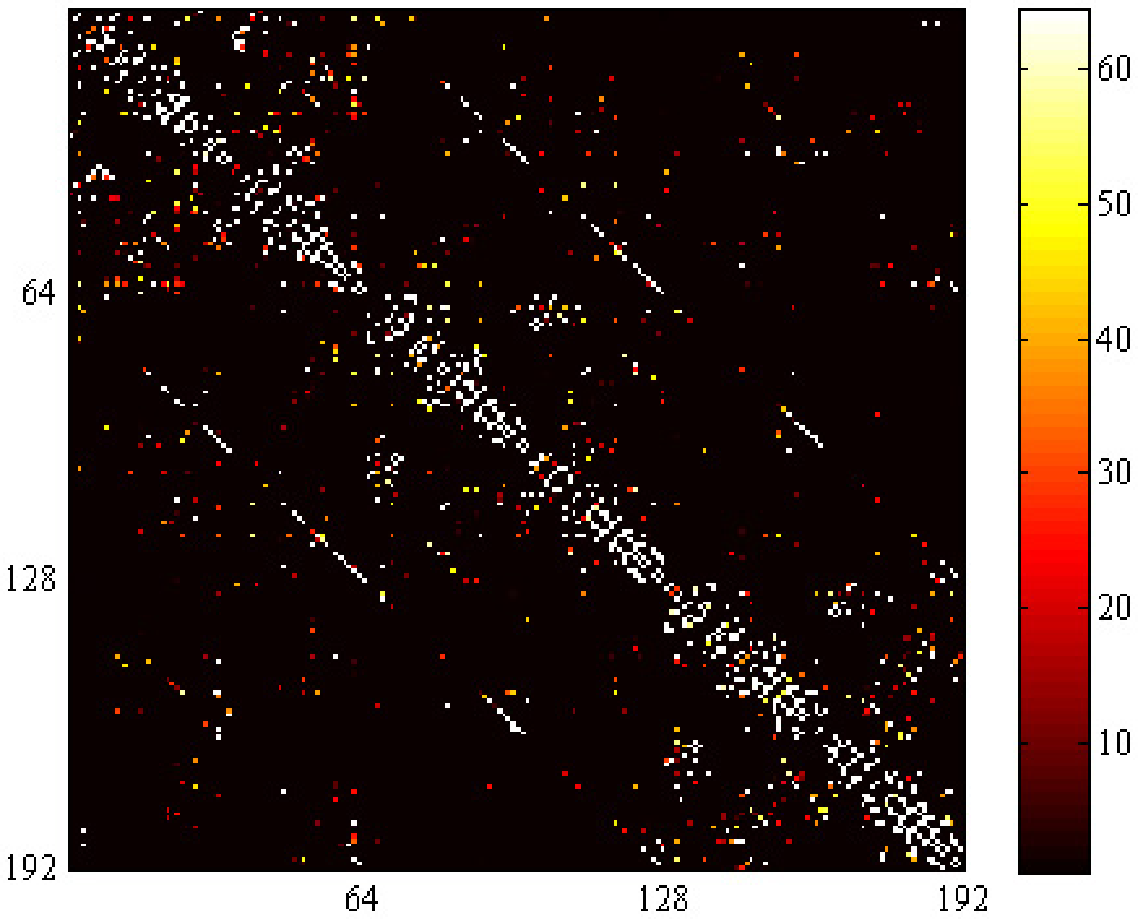}} 
\subfigure[$\Theta^{(1)}$ (27.60\%)]{\includegraphics[clip=true,trim=20 10 30 15,width=0.50\columnwidth]{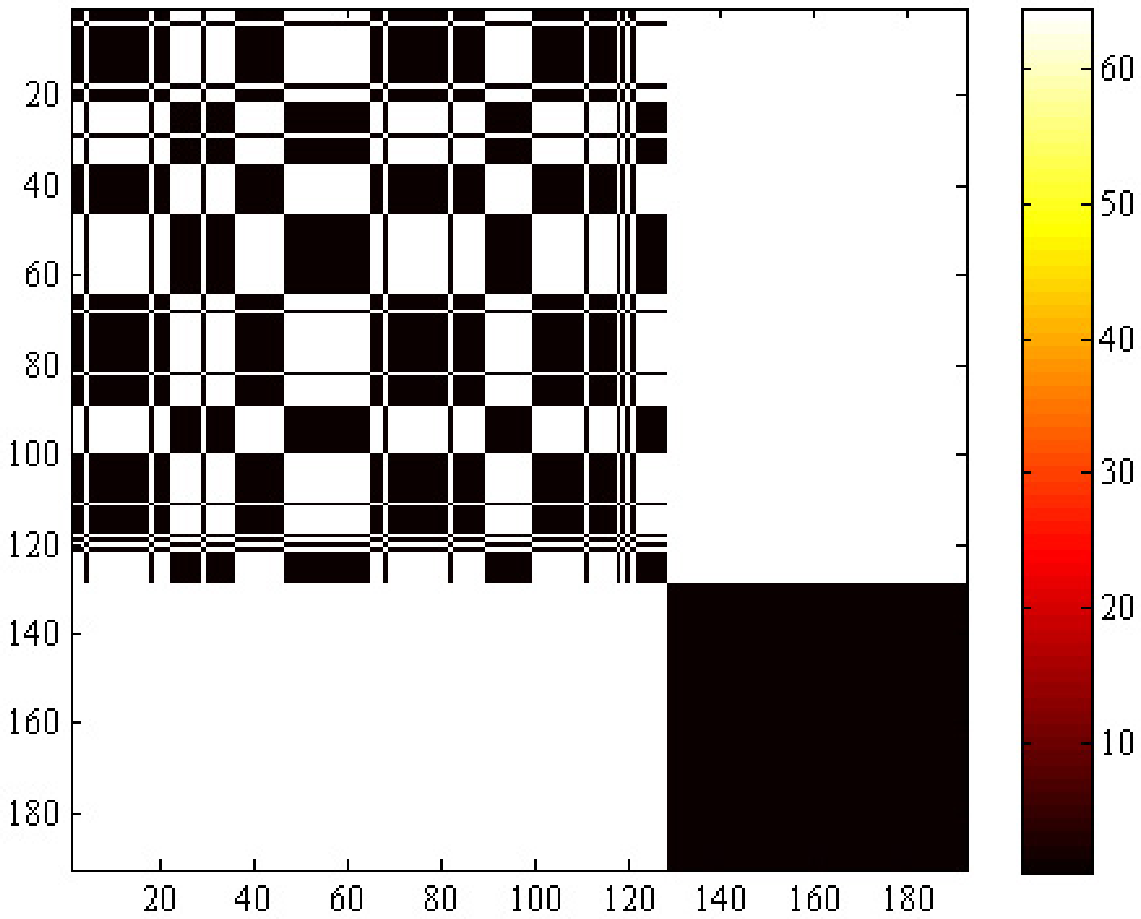}}
\subfigure[$C^{(3)}$]{\includegraphics[clip=true,trim=20 10 30 15,width=0.50\columnwidth]{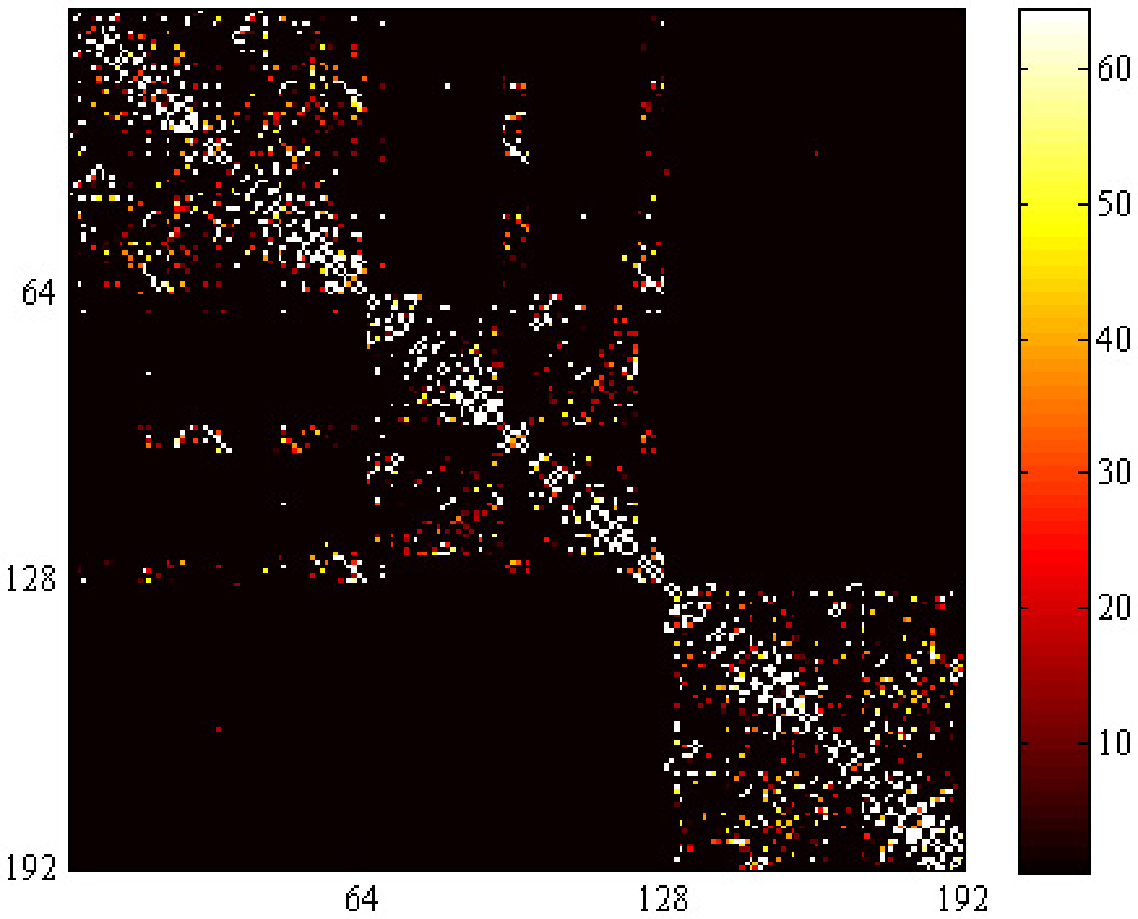}} 
\subfigure[$\Theta^{(3)}$ (6.77\%)]{\includegraphics[clip=true,trim=20 10 30 15,width=0.50\columnwidth]{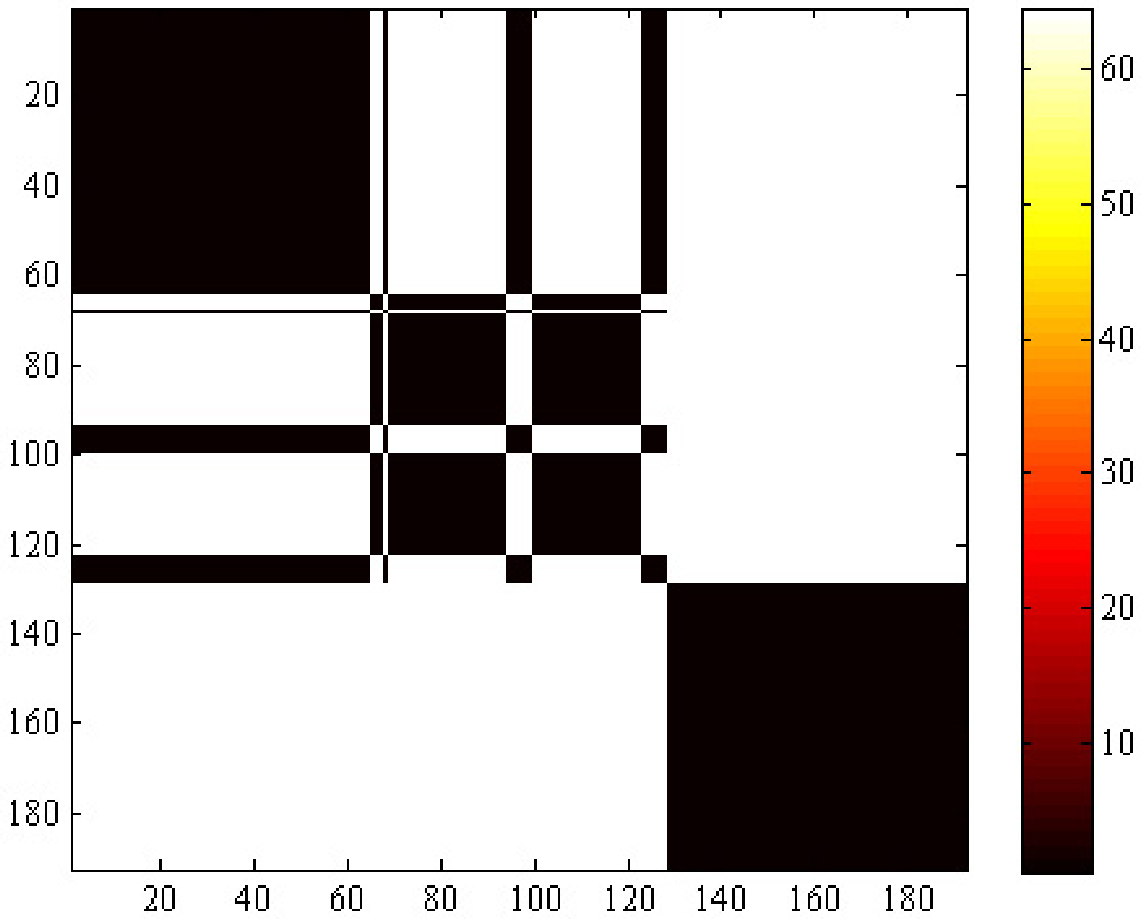}}
\caption{Visualization of representation matrix $C^{(t)}$ and structure matrix $\Theta^{(t)}$ in the first and third iterations of hard \s3c.
Note that the first iteration of hard \s3c is the same as that of SSC and hence the images of $C^{(1)}$ and $\Theta^{(1)}$ in panels (a) and (b) are the representation matrix and the structure matrix of SSC. The structure matrix $\Theta^{(t)}$ is used to re-weight the computation of the representation matrix $C^{(t+1)}$ in the next iteration. The images in panels (c) and (d) are the representation matrix $C^{(t)}$ and structure matrix $\Theta^{(t)}$ of hard \s3c when converged ($t=3$). The percentage numbers in brackets are the corresponding clustering errors. For ease of visualization, we computed the entry-wise absolute value and amplified each entry $|C_{ij}|$ and $\Theta_{ij}$ by a factor of $500$.}
\label{Fig:Z-Theta-visualization-3}
\end{figure*}


\begin{figure*}[tb]
\centering
\subfigure[$C^{(1)}$]{\includegraphics[clip=true,trim=20 10 30 15,width=0.50\columnwidth]{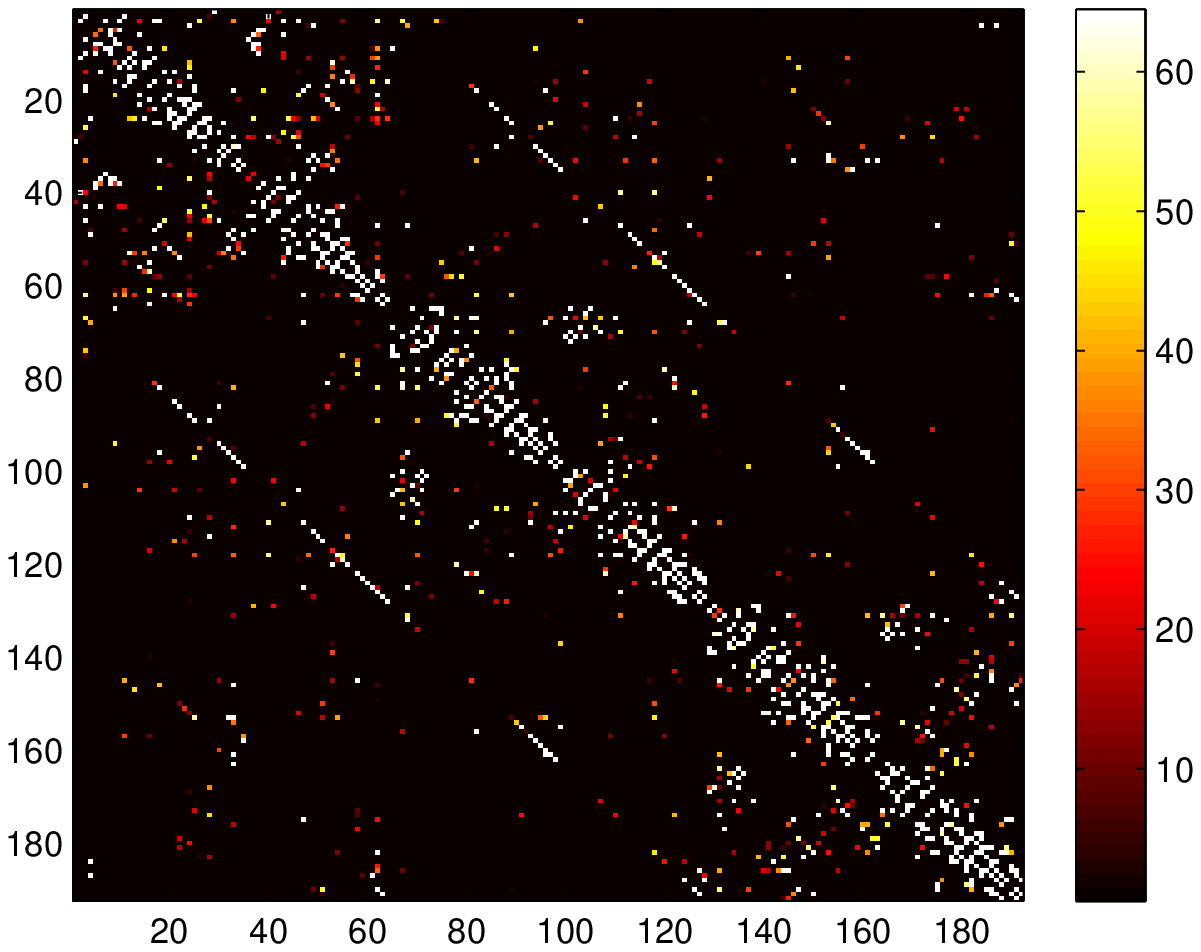}} 
\subfigure[soft $\Theta^{(1)}$ (27.60\%)]{\includegraphics[clip=true,trim=20 10 30 15,width=0.50\columnwidth]{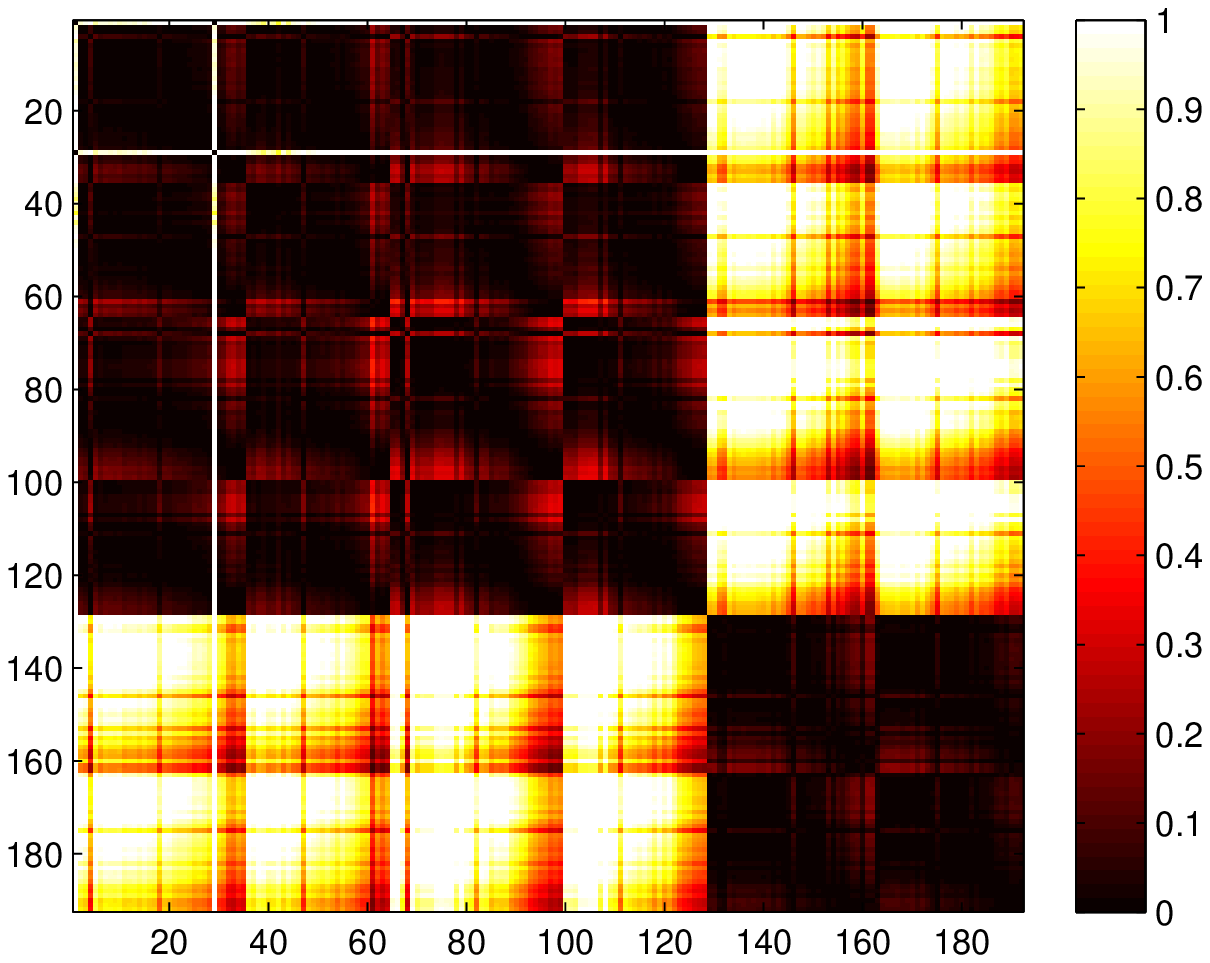}} 
\subfigure[$C^{(4)}$]{\includegraphics[clip=true,trim=20 10 30 15,width=0.50\columnwidth]{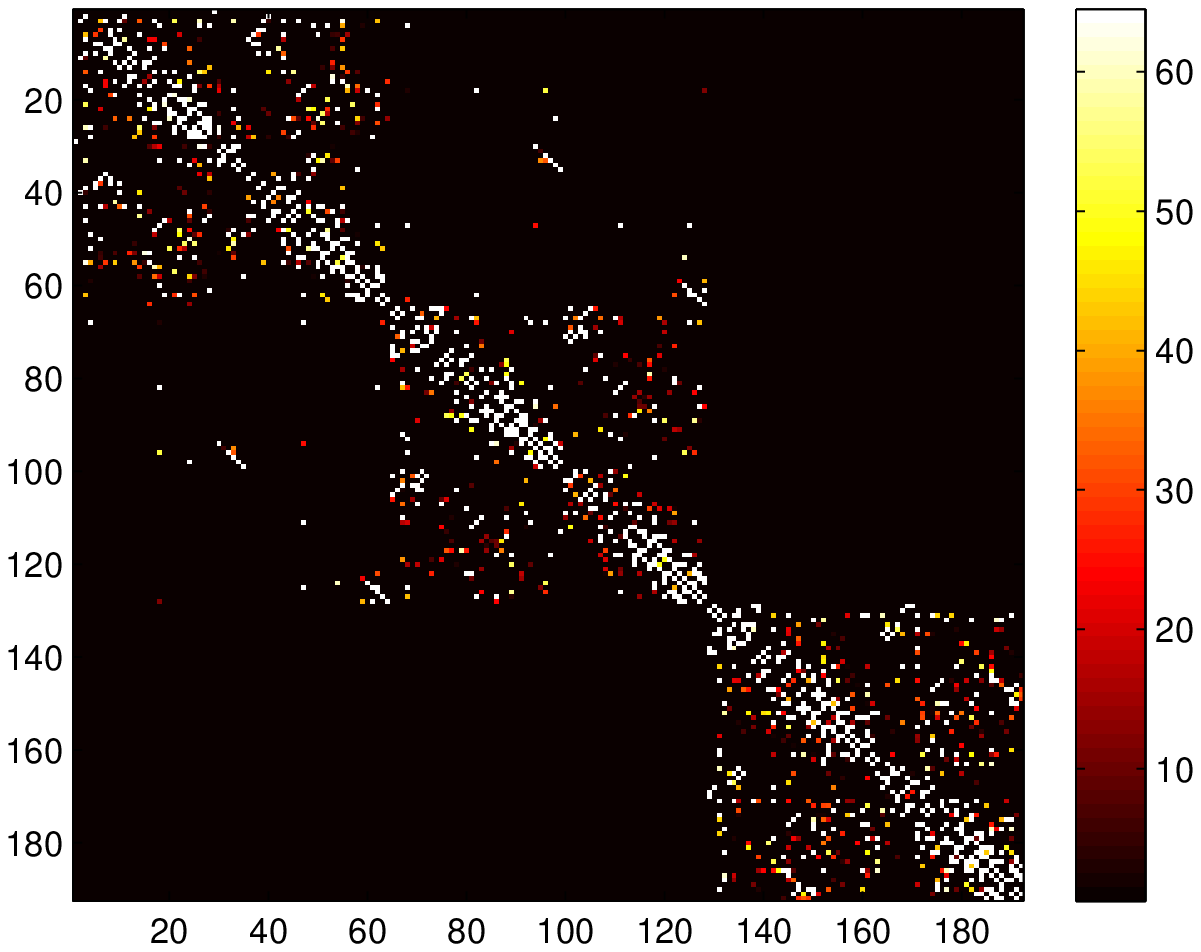}} 
\subfigure[soft $\Theta^{(4)}$ (1.56\%)]{\includegraphics[clip=true,trim=20 10 30 15,width=0.50\columnwidth]{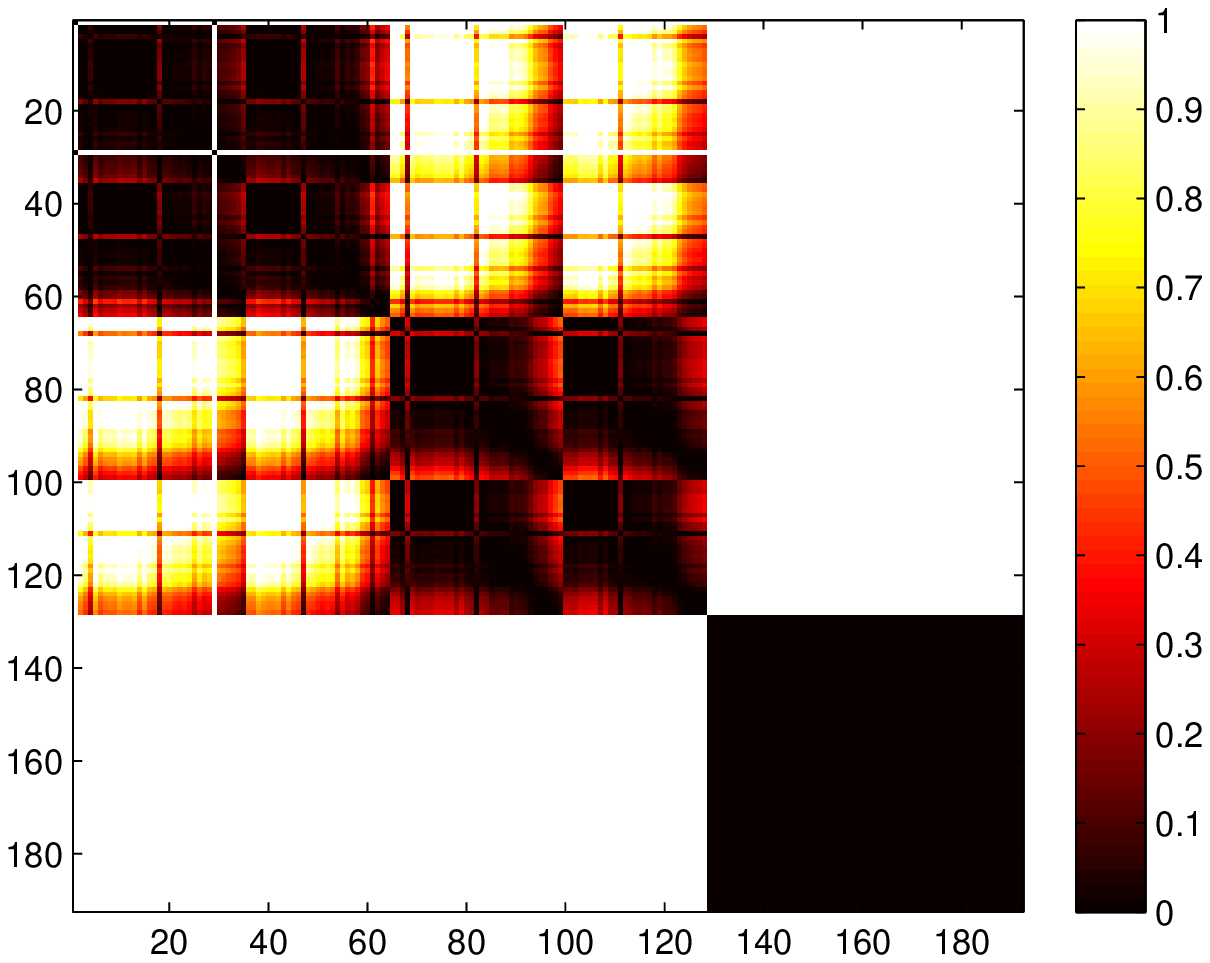}} 
\caption{Visualization of representation matrix $C^{(t)}$ and continuous real-valued structure matrix $\Theta^{(t)}$ in the first and the forth iterations of soft \s3c. The images of $C^{(1)}$ and the soft $\Theta^{(1)}$ in panels (a) and (b) are the representation matrix of SSC and the soft structure matrix computed from the results from SSC. The panels (c) and (d) show the representation matrix $C^{(t)}$ and the soft structure matrix $\Theta^{(t)}$ when converged ($t=4$). The percentage numbers in bracket are the clustering errors. For ease of visualization, we computed the entry-wise absolute value and amplified each entry $|C_{ij}|$ by a factor of $500$.}
\label{Fig:Z-Theta-visualization-3-softS3C}
\end{figure*}

\subsection{Experiments on Extended Yale B Data Set}

Given the face images of multiple subjects acquired under a fixed pose and varying illumination, we consider the problem of clustering the images according to their subjects. It has been shown that, under the Lambertian assumption, the images of a subject with a fixed pose and varying illumination lie close to a linear subspace of dimension 9 \cite{Ho:CVPR03}. Thus, the collection of face images of multiple subjects lie close to a union of 9-dimensional subspaces. In our experiments, we consider the Extended Yale Database B \cite{Kriegman:PAMI01}, which contains 2,414 frontal face images of 38 subjects, with approximately 64 frontal face images per subject taken under different illumination conditions. It should be noted that the Extended Yale B dataset is challenging for subspace clustering due to corruptions in the data caused by specular reflections.

\myparagraph{Data Preprocessing}
In our experiments, we follow the protocol introduced in \cite{Elhamifar:TPAMI13}: a) each image is down-sampled to $48\times 42$ pixels and then vectorized to a 2,016-dimensional data point; 
b) the 38 subjects are then divided into 4 groups -- subjects 1-10, 11-20, 21-30, and 31-38. We perform experiments using all choices of $n \in \{2, 3,5,8,10\}$ subjects in  each of the first three groups and use all choices of $n\in \{2,3,5,8\}$ from the last group.
For \s3c, SSC \cite{Elhamifar:TPAMI13}, LatLRR \cite{Liu:ICCV11}, and LRSC~\cite{Vidal:PRL14}, we use the 2,016-dimensional vectors as inputs. For LatLRR and LRSC, we cite the results reported in \cite{Elhamifar:TPAMI13} and \cite{Vidal:PRL14}, respectively.
For LRR \cite{Liu:ICML10}, LSR \cite{Lu:ECCV12}, and CASS \cite{Lu:ICCV13-TraceLasso}, we use the procedure reported in \cite{Lu:ECCV12}: use standard PCA to reduce the 2,016-dimensional data to $6n$-dimensional data for $n \in \{2,3,5,8,10\}$.

\myparagraph{Performance of \s3c as a Function of $\alpha$} To show the effect of the parameter $\alpha$ on the performance of \s3c, we conduct experiments on all 163 
choices of $n=2$ subjects on the Extended Yale B data set with $\alpha \in \{10^{-4}, 10^{-3}, 10^{-2}, 0.1, 0.5, 1, 2, 5, 10, 15, 20\}$. For each $\alpha$, we calculate the average ERR, CONN, and SPR over the 163 choices and present them as a function of $\alpha$. Experimental results are shown in Fig.~\ref{fig:ERR-CONN-SRR-ZQnorm-vs-alpha}. We can observe that:
a) Soft \s3c can yield more accurate results when $\alpha > 0.1$;
b) Using a middle value for $\alpha$, CONN can be improved; 
c) By setting $\alpha$ not too small, SPR can be improved. 
Thus, a reasonable range for $\alpha$ in both algorithms is the interval $[10^{-3}, 2]$. By default, we use $\alpha = 0.1$ for hard \s3c and $\alpha = 1$ for soft \s3c in the face clustering experiments.

\begin{figure*}[tb]
\centering
\subfigure[]{\includegraphics[clip=true,trim=0 0 25 0,width=0.45\columnwidth]{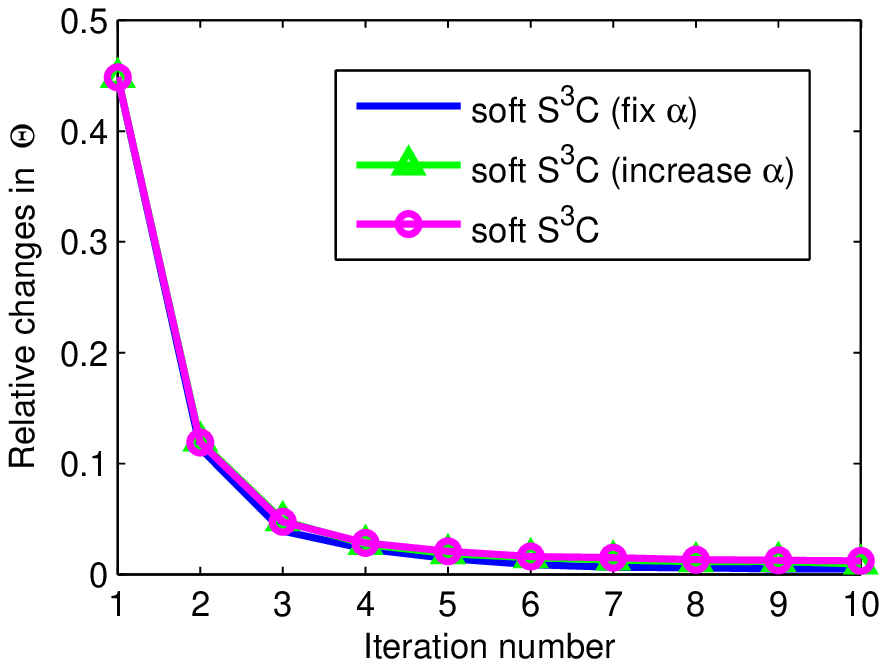}}
\subfigure[]{\includegraphics[clip=true,trim=0 0 25 0,width=0.45\columnwidth]{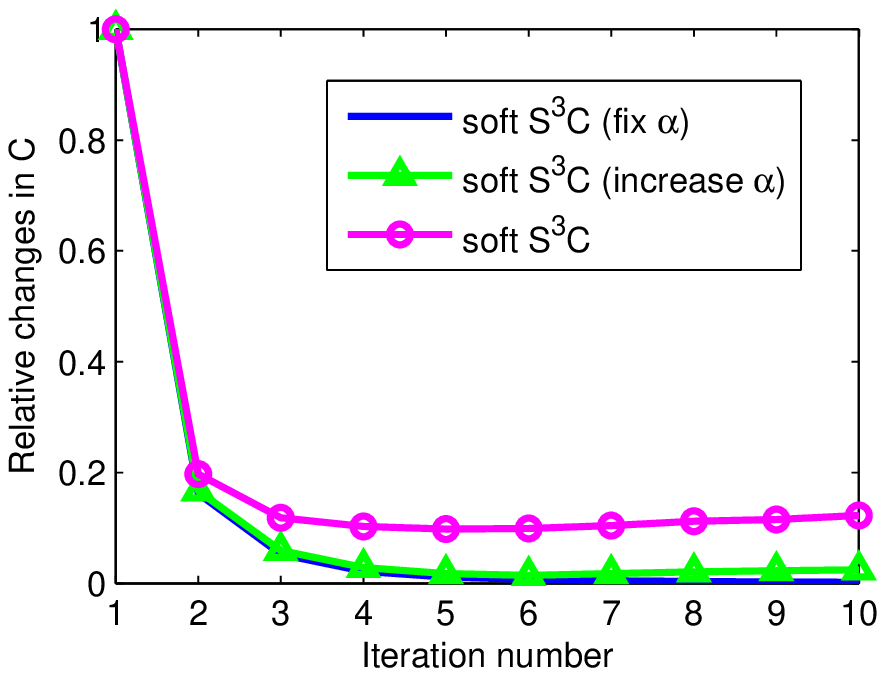}}
\subfigure[]{\includegraphics[clip=true,trim=0 0 25 0,width=0.45\columnwidth]{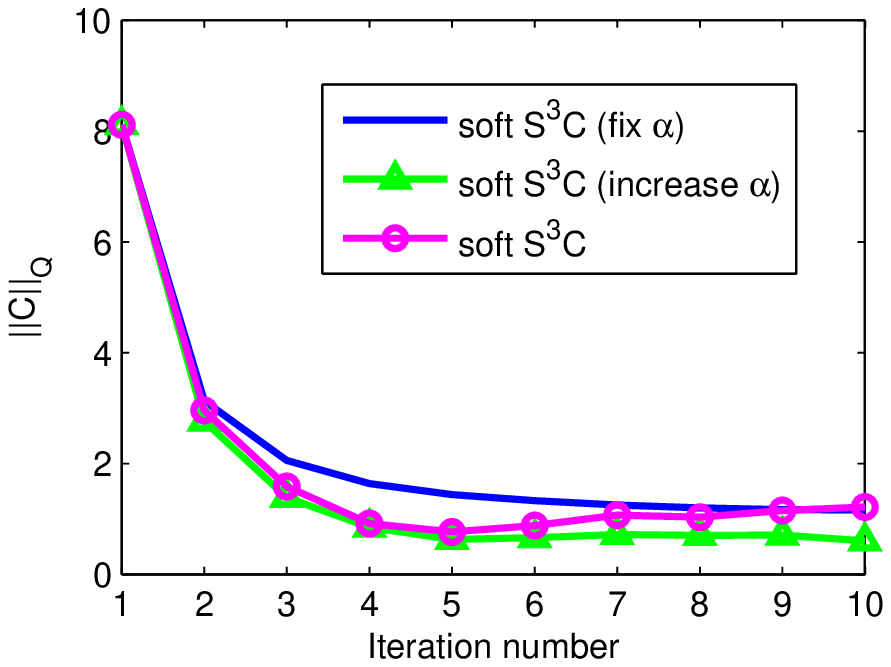}}
\subfigure[]{\includegraphics[clip=true,trim=0 0 25 0,width=0.45\columnwidth]{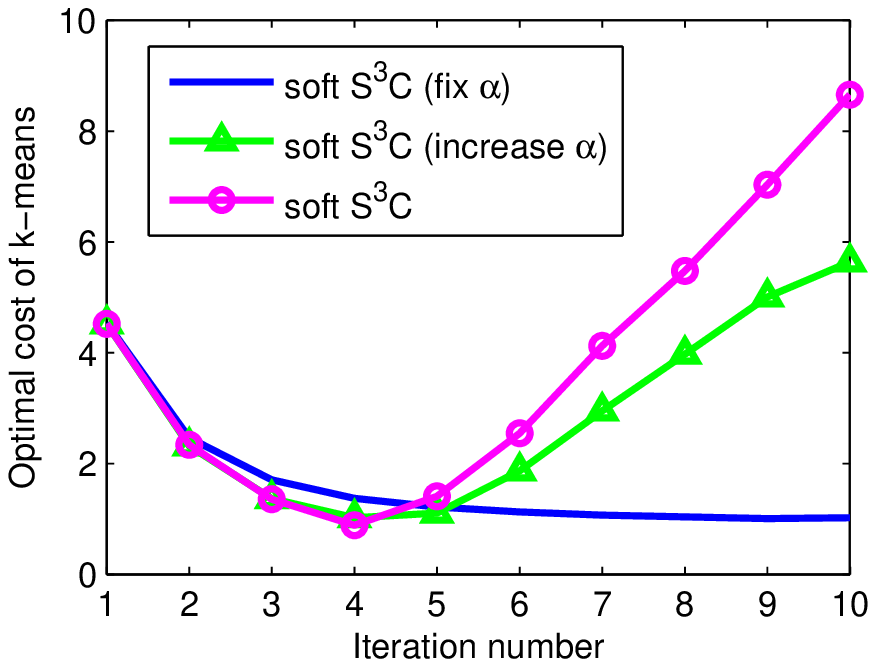}}
\caption{Experimental evaluation on the convergence behavior of soft \s3c with three types of $\alpha$ settings. Panels (a) and (b) show relative changes of subspace structure matrix $\Theta$ and coefficient matrix $C$ in two consecutive iterations, respectively. Panels (c) and (d) show subspace structured norm of $C$ w.r.t. $Q$ and the optimal cost of $k$-means in each iteration, respectively.}
\label{fig:Rel-Theta-C-Z-Q-norm-Cost-kmeans-vs-iter-goUpDown-alpha-1p0}
\end{figure*}

\begin{figure}[tb]
\centering
\subfigure[ERR]{\includegraphics[clip=true,trim=0 0 15 10,width=0.325\columnwidth]{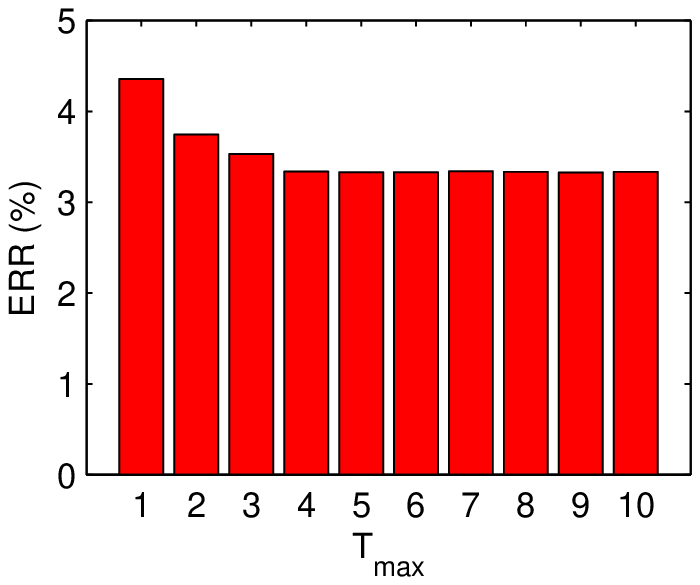}}
\subfigure[CONN]{\includegraphics[clip=true,trim=0 0 15 10,width=0.325\columnwidth]{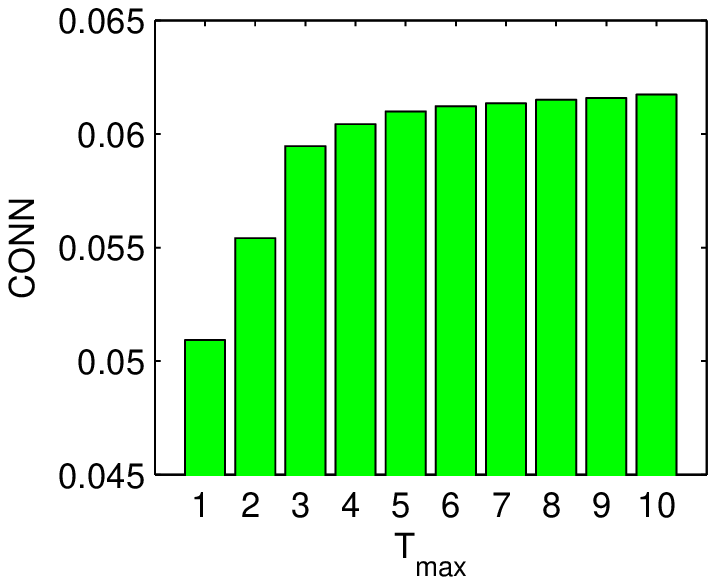}}
\subfigure[SRR]{\includegraphics[clip=true,trim=0 0 15 10,width=0.325\columnwidth]{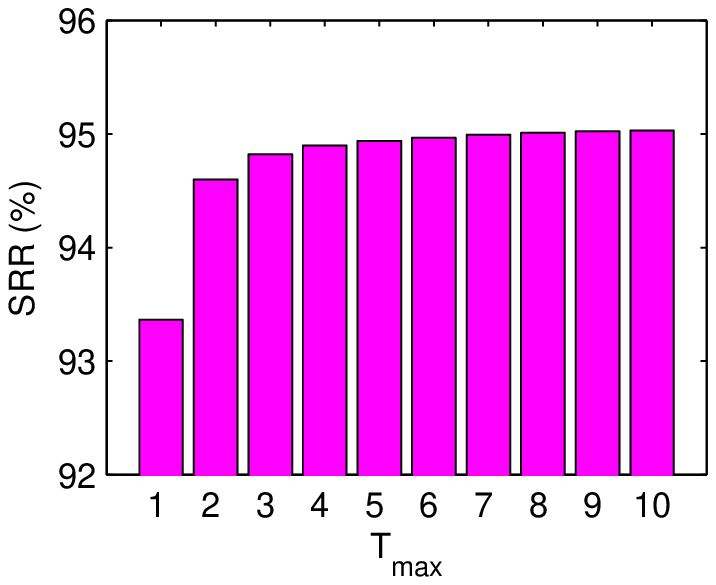}}
\\
\subfigure[ERR]{\includegraphics[clip=true,trim=0 0 15 5,width=0.325\columnwidth]{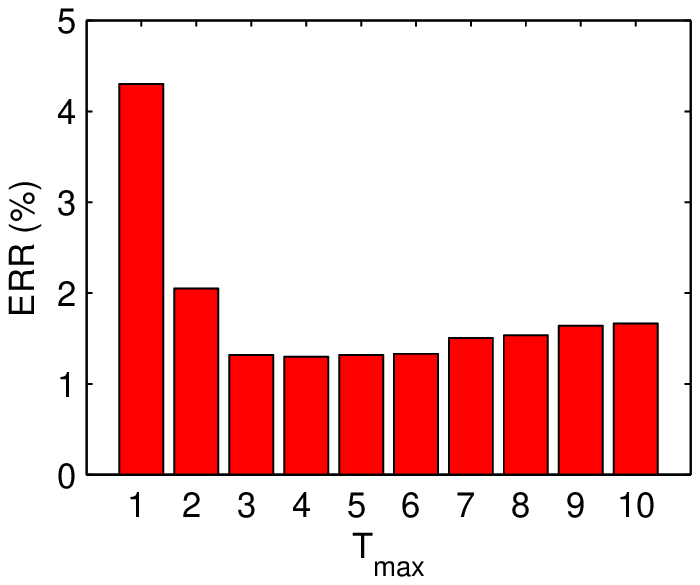}}
\subfigure[CONN]{\includegraphics[clip=true,trim=0 0 15 5,width=0.325\columnwidth]{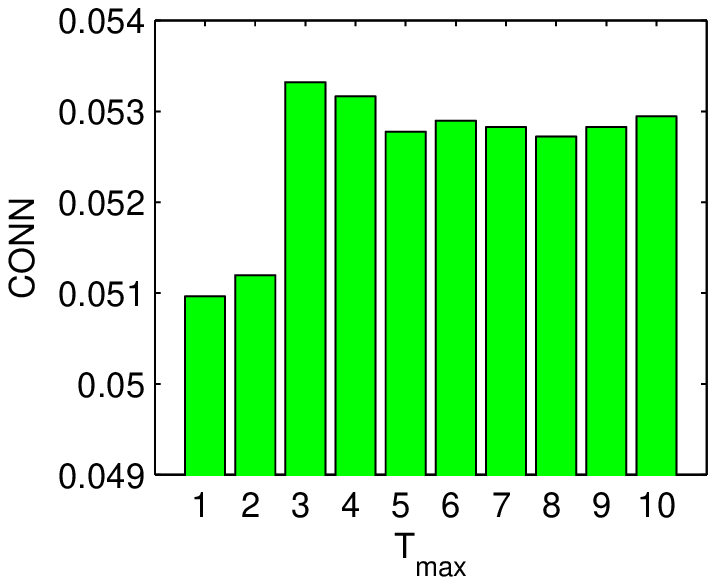}}
\subfigure[SPR]{\includegraphics[clip=true,trim=0 0 15 5,width=0.325\columnwidth]{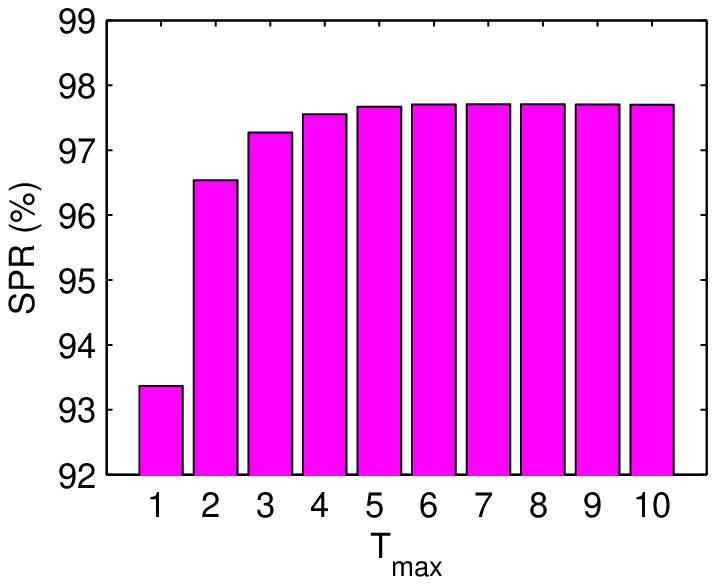}}
\caption{Evaluation on effect of subspace sparse $\ell_1$ norm during iterations of \s3c. Top row: hard \s3c; Bottom row: soft \s3c}
\label{fig:ERR-SRR-CONN-ZQ-norm-vs-iter-fixed-alpha}
\end{figure}

\begin{table*}[!tb]
\caption{{\mcb Clustering Errors on Extended Yale B. The best results are in bold font. In \s3c$^\dag$ we fix $\alpha$ as $0.1$ for the hard version and $1.0$ for the soft version, respectively; whereas in \s3c$^\ddag$ we use $\alpha \leftarrow \nu \alpha$.}} 
\centering
\begin{tabular}{|c|cc|cc|cc|cc|cc|}
\hline
No. subjects     & 2       &         & 3      &       &5        &         & 8      &          & 10     &       \\
ERR (\%)         & Average & Median  & Average& Median& Average & Median  & Average& Median   & Average& Median\\
\hline
\hline
LRR
                       & 6.74$\pm$4.22    & 7.03    & 9.30$\pm$3.63   &9.90   &13.94$\pm$3.36     & 14.38   & 25.61$\pm$5.08   & 24.80    & 29.53$\pm$4.32    & 30.00 \\ 
LSR1
                       & 6.72$\pm$4.16    & 7.03    & 9.25$\pm$3.64    &9.90    &13.87$\pm$3.40      &14.22      & 25.98$\pm$5.48   & 25.10     & 28.33$\pm$5.65    &30.00  \\
LSR2
                       & 6.74$\pm$4.22    & 7.03    & 9.29$\pm$3.64    &9.90    &13.91$\pm$3.40      &14.38      & 25.52$\pm$5.47   & 24.80     & 30.73$\pm$3.29    &33.59  \\
CASS        &10.95$\pm$12.22 &6.25   & 13.94$\pm$14.22    &7.81      & \ 21.25$\pm$13.70    &18.91      & 29.58$\pm$5.66    & 29.20      & \ 32.08$\pm$11.59    & 35.31 \\
LRSC \cite{Vidal:PRL14}& 3.15    & 2.34    & 4.71    &4.17    &13.06     &8.44      & 26.83   & 28.71     &35.89     & 34.84 \\
BDSSC \cite{Feng:CVPR14} & 3.90    & -       & 17.70   & -      &27.50     & -        & 33.20   & -         &39.53     & -     \\
BDLRR \cite{Feng:CVPR14} & 3.91    & -       & 10.02   & -      &12.97     & -        & 27.70   & -         &30.84     & -     \\
LatLRR \cite{Liu:ICCV11} & 2.54    & 0.78    & 4.21    & 2.60   &6.90      & 5.63     & 14.34   &10.06      &22.92     & 23.59       \\
TSC \cite{Heckel:TIT15} & 8.06    & -   & 9.00   &  - & 10.14     & -    & 12.58   & -     & 17.86     & -         \\
OMP \cite{Dyer:JMLR13}    & 4.45    & -    & 6.35   &  - &8.93     & -    & 12.90    & -     & 9.82    & -         \\
NSN\cite{Park:NIPS14}   & 1.71    & \textbf{0.00}    & 3.63   &  - &5.81     & -    & 8.46    & -     & 9.82     & -         \\
SSC               & 1.87$\pm$6.39    & \textbf{0.00}    & 3.35$\pm$7.02   &  0.78 &4.32$\pm$4.60      & 2.81    & 5.99$\pm$4.13    & 4.49     & 7.29$\pm$4.28     & 5.47         \\
\hline
\text{\!hard \s3c$^\dag$} \cite{Li:CVPR15}  & \text{1.43$\pm$5.62}    & \textbf{0.00}    & \text{3.09$\pm$6.90}   &  \textbf{0.52} &\text{4.08$\pm$4.85}     & \text{2.19}    &\text{4.84$\pm$3.07}   & \text{4.10}     & \text{6.09$\pm$3.38} & \text{5.16} \\
\text{hard \s3c$^\ddag$} \cite{Li:CVPR15}  & \text{1.40$\pm$5.61}    & \textbf{0.00}    & \text{3.08$\pm$7.01}   &  \textbf{0.52} &\text{3.83$\pm$4.75}     & \text{1.88}    &\text{4.45$\pm$3.11}   & \text{3.52}     & \text{5.42$\pm$3.14} & \text{4.53} \\
\text{\!\!\!\!hard \s3c} \cite{Li:CVPR15}  & \text{1.27$\pm$5.54}    & \textbf{0.00}    & \text{2.71$\pm$6.80}   &  \textbf{0.52} &\text{3.41$\pm$4.88}     & \text{1.25}    &\text{4.15$\pm$3.22}   & \text{2.93}     & \text{5.16$\pm$4.30} & \text{4.22} \\
%
%
%
\hline
\textbf{\!\!soft $\bf S^3$C$^\dag$}  & \textbf{0.52$\pm$1.25}    & \textbf{0.00}    & \textbf{0.89$\pm$1.15}   &  \textbf{0.52} &\textbf{1.51$\pm$1.07}     & \textbf{1.25}    &\textbf{2.31$\pm$0.97}   & \textbf{2.25}     & \textbf{2.81$\pm$0.68} & \textbf{2.50} \\


\textbf{\!soft $\bf S^3$C$^\ddag$}  & \textbf{0.51$\pm$1.18}    & \textbf{0.00}    & \textbf{0.94$\pm$1.21}   &  \textbf{0.52}   & \textbf{1.65$\pm$1.38}     & \textbf{1.56}    &\textbf{2.54$\pm$1.44}   & \textbf{2.34}     & \textbf{2.92$\pm$0.70} & \textbf{2.97} \\

\textbf{\!\!\!\!\!\! soft $\bf S^3$C}  & \textbf{0.76$\pm$3.90}    & \textbf{0.00}    & \textbf{0.82$\pm$1.14}   &\textbf{0.52} &{\textbf{1.32$\pm$0.99}}     & \textbf{1.25}    &{\textbf{2.14$\pm$1.05}}   & \textbf{1.95}     &{\textbf{2.40$\pm$1.10}} & \textbf{2.50} \\
\hline
\end{tabular}
\label{table:extendedYaleB-SSC-vs-SSSC}
\end{table*}

\myparagraph{Data Visualization: Hard $\Theta$ vs. Soft $\Theta$} To show the effect of using the subspace structured $\ell_1$ norm in \s3c, and especially the differences in using the hard and the soft $\Theta$ intuitively, we apply our \s3c algorithms to a subset of the Extended Yale B and visualize the representation matrix $C^{(t)}$ and the subspace structure matrix $\Theta^{(t)}$, which are taken in the $t$-th iteration. The visualization results are shown in Fig.~\ref{Fig:Z-Theta-visualization-3} and Fig.~\ref{Fig:Z-Theta-visualization-3-softS3C}.

We observe from Fig.~\ref{Fig:Z-Theta-visualization-3} (a) that the structured sparse representation matrix $C^{(1)}$---which is the same as that of SSC---is not block-diagonal and leads to a degenerate clustering result as shown by the ``hard'' subspace structure matrix $\Theta^{(1)}$ in Fig.~\ref{Fig:Z-Theta-visualization-3} (b). In the third iteration ($t=3$), hard \s3c yields a much better representation matrix $C^{(3)}$, as shown in Fig.~\ref{Fig:Z-Theta-visualization-3} (c), and hence produces a significantly improved clustering result as shown by the binary subspace structure matrix $\Theta^{(3)}$ in Fig.~\ref{Fig:Z-Theta-visualization-3} (d). While hard \s3c in this case did not yield a perfect block-diagonal representation matrix, the improvements still reduce the clustering error from 27.60\% to 6.77\%.

Compared to the binary ``hard'' structure matrix $\Theta$, as shown in Fig.~\ref{Fig:Z-Theta-visualization-3} (b) and (d), the continuous real-valued ``soft'' structure matrix $\Theta$, as shown in Fig.~\ref{Fig:Z-Theta-visualization-3-softS3C} (c) and (d), carries more detailed information about the confidence or uncertainty in the previous clustering results. Although it did not converge to an exact block diagonal matrix, the improved representation matrices $C^{(t)}$'s give more accurate clustering results. In this case, the soft \s3c reduces the clustering error from 27.60\% to 1.56\%.

\myparagraph{Convergence Behaviors}
To show the convergence behaviors of Algorithm \ref{alg:hard-S3C}, we perform soft \s3c on the Extended Yale B data set for all choices of $n=5$ with the parameter set to $\alpha=1$. We calculate the relative changes of subspace structure matrix $\Theta$ as defined in \eqref{eq:Alg2-converge-stopping-rules-I}, the relative changes of coefficients matrix $C_\ast$ as in \eqref{eq:Alg2-converge-stopping-rules-II}, and the subspace structured norm of $C$ with respect to segmentation matrix $Q$ (\ie, $\|C\|_Q$), and the optimal cost of $k$-means in spectral clustering, as a function of iteration number, respectively. Experimental results are presented in Fig.~\ref{fig:Rel-Theta-C-Z-Q-norm-Cost-kmeans-vs-iter-goUpDown-alpha-1p0}. As it can be observed from panels (a)-(c), the iterations in Algorithm \ref{alg:hard-S3C} make coefficient matrix $C$ and subspace structure matrix $\Theta$ as consistent as possible. In addition, the ``elbow'' shape tendency in panels (c) and (d) suggests that the subspace structured norm $\|C\|_Q$ and the optimal cost of $k$-means in spectral clustering can be used to perform model selection in \s3c.

\begin{table*}[t]
\caption{{\mcb Motion Segmentation Errors on Hopkins 155. The best results are in bold font. ``h-'': hard. ``s-'': soft.}}
{\vspace{-5pt}}
\begin{center}
\scriptsize
\begin{tabular}{|c c| c| c|c|c| c|  c |c || c| c| c | c| c| c | c|}
\hline
Methods  &    &LSA      &LRR   &BDLRR   & LSR1 & LSR2  &BDSSC  &SSC   &h-\s3c$^\dag$ & h-\s3c$^\ddag$  &h-\s3c &s-\s3c$^\dag$ &s-\s3c$^\ddag$ & s-\s3c      \\
\hline
\hline
2 motions&Ave.&3.27     &3.76      &3.70  &2.20        &2.22     &2.29       &1.95   &1.68 &1.73 &1.73 &\textbf{1.60}  &1.64  &1.65 \\
ERR(\%)  &Med.&0.55 &\textbf{0.00}&\textbf{0.00}&\textbf{0.00}&\textbf{0.00}&\textbf{0.00}&\textbf{0.00}&\textbf{0.00}&\textbf{0.00}&\textbf{0.00} &\textbf{0.00}  &\textbf{0.00} &\textbf{0.00} \\ 
         &Std.&8.41     &7.73     &10.31 &\textbf{5.48}&5.73     &7.75       &7.19   &6.06 &6.25 &6.34    &5.93 &6.15  &6.14 \\
\hline
3 motions&Ave.&9.15     &9.92     &6.49  &7.13         &7.18     &4.95       &4.94   &5.26 &5.29 &5.50    &4.27 &\textbf{4.11}  &4.27 \\
ERR(\%)  &Med.&1.66     &1.42     &1.20  &2.40         &2.40     &0.91       &0.89   &0.81 &0.81 &0.81    &0.73 &0.73  &\textbf{0.61} \\
         &Std.&14.58    &11.33    &12.32 &8.96         &8.86     &9.72       &9.91   &10.35&10.36&10.27   &8.97 &\textbf{8.82}  &8.89 \\
\hline
Total    &Ave.&4.60     &5.15     &4.33  &3.31   &3.34           &2.89       &2.63   &2.49 &2.53 &2.58 &\textbf{2.20} &\textbf{2.20}   &2.24 \\
ERR(\%)  &Med.&0.69&\textbf{0.00}&\textbf{0.00}&0.22&0.23&\textbf{0.00}&\textbf{0.00}&\textbf{0.00}&\textbf{0.00}& \textbf{0.00} &\textbf{0.00} &\textbf{0.00} &\textbf{0.00} \\
         &Std.&10.37    &9.07  &10.82&\textbf{6.72}&6.86           &8.28        &7.95  &7.37 &7.49 &7.54  &6.80  &6.89  &6.92 \\
\hline
\end{tabular}
\end{center}
\label{table:hopkins155-SSC-vs-SSSC}
\end{table*}

\myparagraph{Effect of Using the Subspace Structured $\ell_1$ Norm}
To evaluate the effect of using the subspace structured $\ell_1$ norm, we conduct experiments with 
\s3c on the Extended Yale B data set for all cases of $n=5$, with the parameter set to 
$\alpha=0.1$ for hard \s3c and $\alpha=1$ for soft \s3c. We calculate the average ERR, CONN, and SPR as a function of the maximum iteration number. Experimental results are presented in Fig.~\ref{fig:ERR-SRR-CONN-ZQ-norm-vs-iter-fixed-alpha}. As it can be observed, the subspace-preserving property of the coefficients and the connectivity of the induced affinity matrix are improved during iterations. Note that the clustering errors are reduced remarkably in 2 or 3 iterations. This observation suggests that it is enough to stop our proposed \s3c algorithms in $2$ or $3$ iterations.

\myparagraph{Comparison to State-of-the-art} We compare the clustering errors of our proposed two \s3c algorithms with SSC \cite{Elhamifar:TPAMI13}, LRR \cite{Liu:ICML10}, LatLRR \cite{Liu:ICCV11}, LRSC~\cite{Vidal:PRL14}, and other recently proposed algorithms, including LSR \cite{Lu:ECCV12}, CASS \cite{Lu:ICCV13-TraceLasso}, and BDSSC \cite{Feng:CVPR14}, BDLRR \cite{Feng:CVPR14}, TSC \cite{Heckel:TIT15}, OMP \cite{Dyer:JMLR13}, and NSN\cite{Park:NIPS14}. For BDSSC, BDLRR, TSC, OMP and NSN, we directly cite their best results. 
{\mcb In \s3c, $\lambda$ is set the same as SSC in which $\lambda_0 =20$.}
The full experimental results are presented in Table~\ref{table:extendedYaleB-SSC-vs-SSSC}.
As it can be observed, the hard \s3c algorithm \cite{Li:CVPR15} outperforms all other algorithms in clustering errors across all experimental conditions, and the soft \s3c algorithm further reduce the clustering error significantly.

Moreover, we also conduct experiments on the whole data set for all 38 subjects. In this case, the average clustering error of SSC over 10 trials is
$34.09\%$; 
whereas the average clustering errors of hard \s3c and soft \s3c are $27.04\%$ 
and $19.82\%$, 
respectively.
Note that 
the clustering error of soft \s3c can be further reduced to $15.50\%$ if we use $\alpha=0.5$.


\begin{figure}
\centering
\includegraphics[width=0.325\columnwidth]{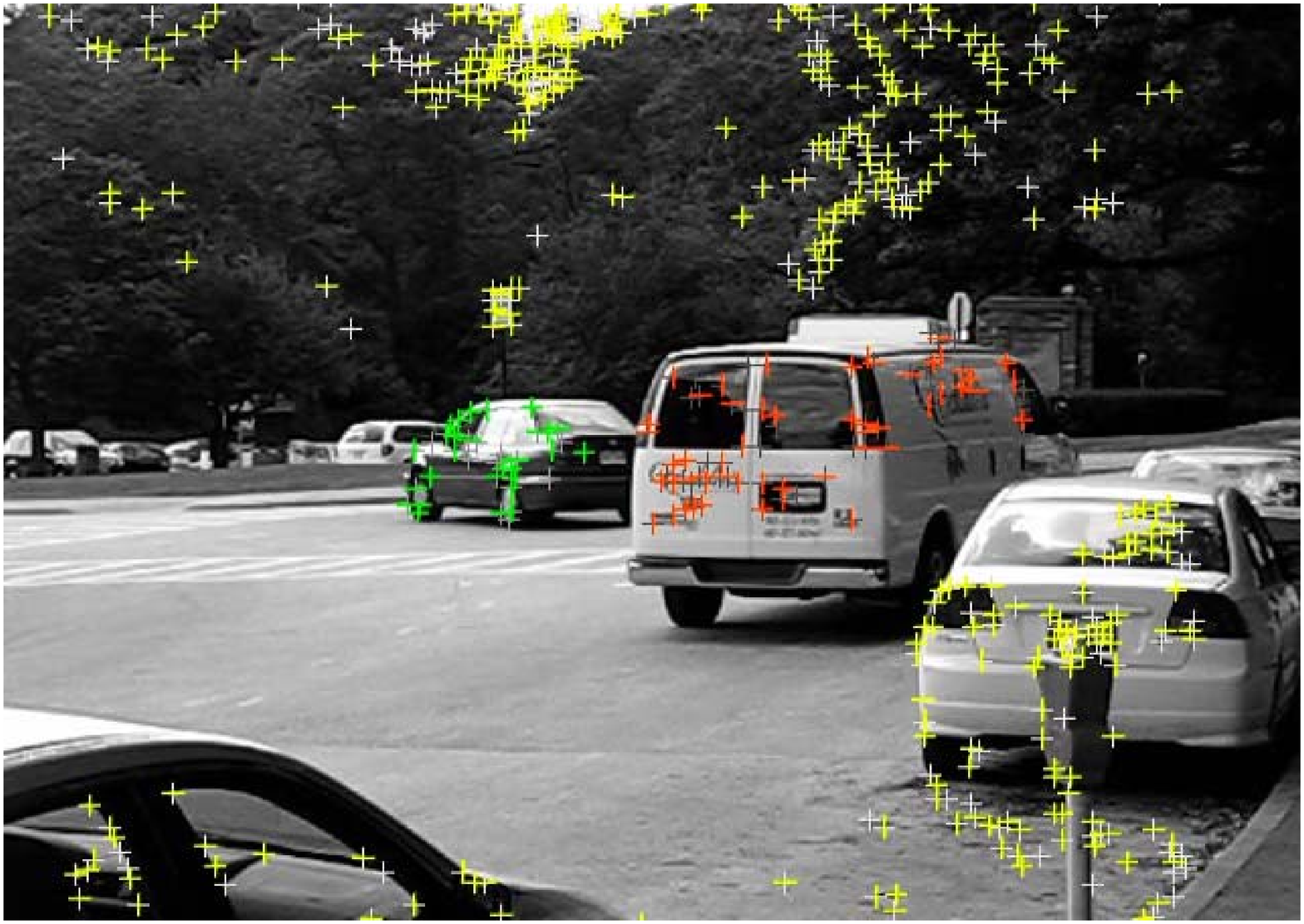}
\includegraphics[width=0.325\columnwidth]{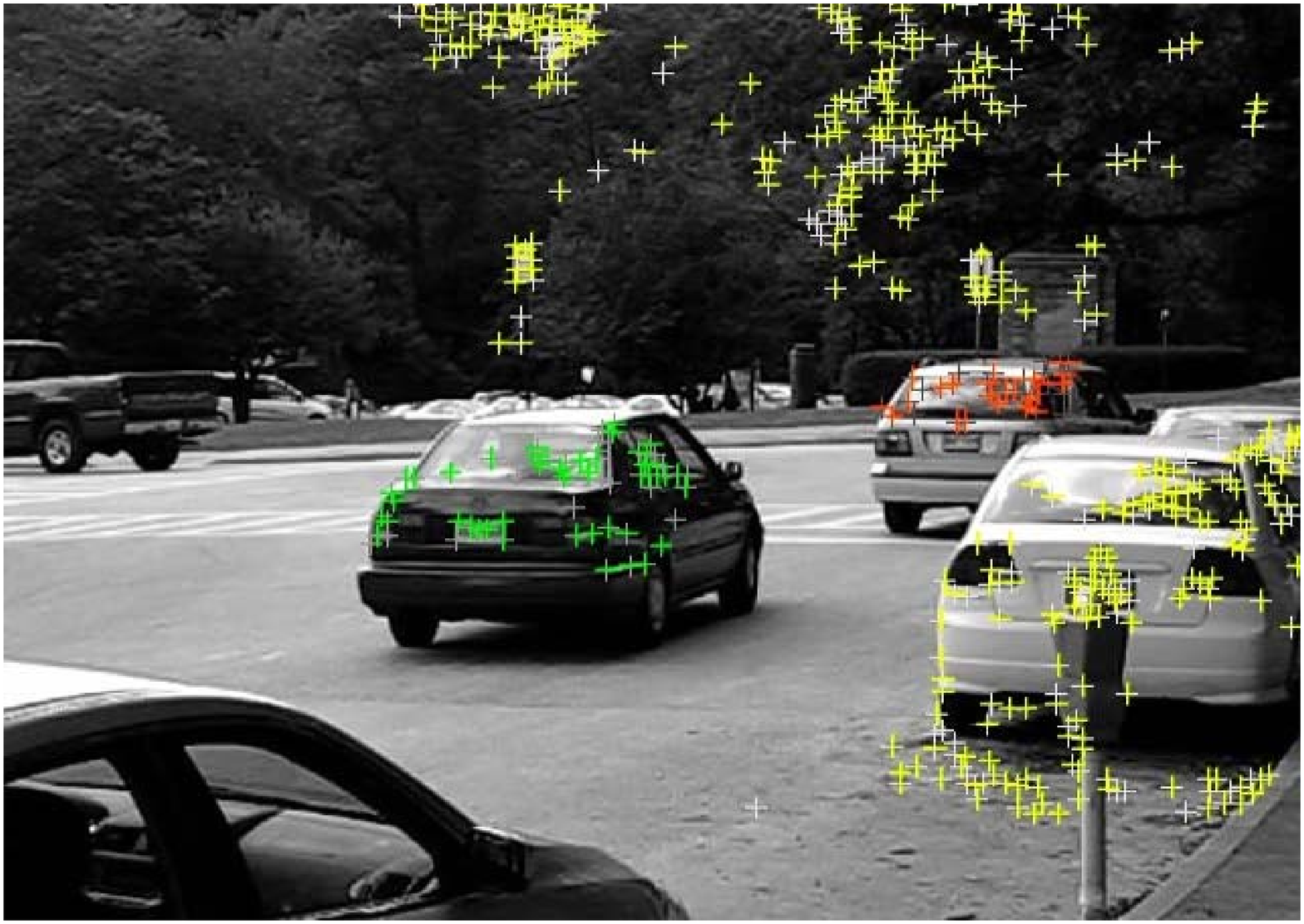}
\includegraphics[width=0.325\columnwidth]{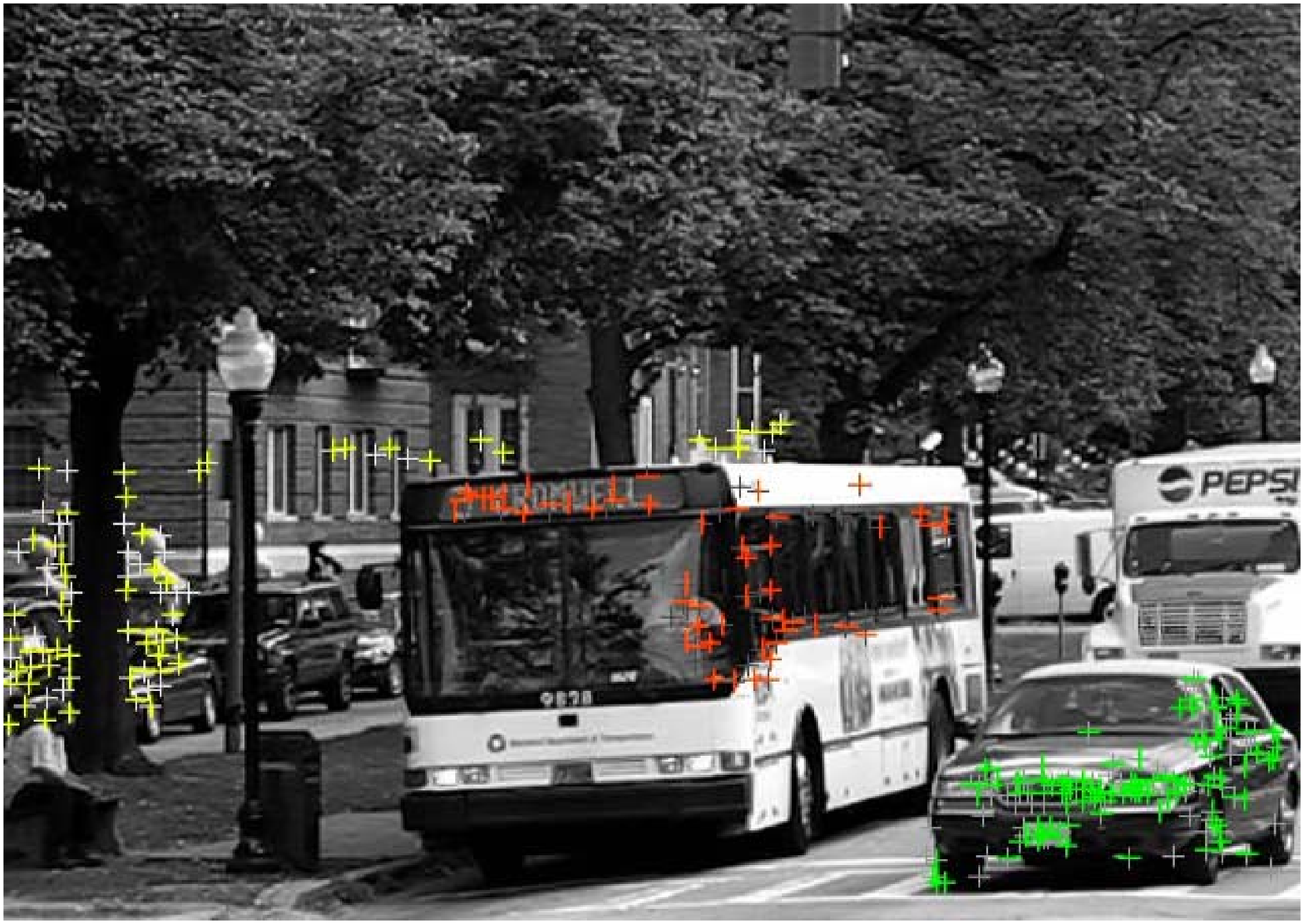}
\caption{Example frames from videos in the Hopkins 155~\cite{Tron:CVPR07}.}
\label{fig_MoSeg12}
{\vspace{-5pt}}
\end{figure}

\subsection{Experiments on Hopkins 155 Database}

Motion segmentation refers to the problem of segmenting a video sequence with multiple rigidly moving objects into multiple spatiotemporal regions that correspond to the different motions in the scene (see Fig. \ref{fig_MoSeg12}).
This problem is often solved by first extracting and tracking the spatial positions of a set of $N$ feature points $\textbf{x}_{fi} \in \RR ^2$ through each frame $f=1,\dots,F$ of the video, and then clustering these feature 
point trajectories according to each one of the motions. Under the affine projection model, a feature point trajectory is formed by stacking the feature points $\textbf{x}_{fi}$ in the video as $\textbf{y}_i \doteq [\textbf{x}^\top_{1i}, \textbf{x}^\top_{2i}, \cdots, \textbf{x}^\top_{Fi} ]^\top \in \RR ^{2F}$. Since the trajectories associated with a single rigid motion lie in an affine subspace of $\RR ^{2F}$ of dimension at most $3$ 
\cite{Tomasi:IJCV92}, the trajectories of $n$ rigid motions lie in a union of $n$ low-dimensional subspaces of $\RR ^{2F}$. Therefore, the multi-view affine motion segmentation problem reduces to the subspace clustering problem.

\begin{table*}[tb]
\caption{{\mcb Clustering error (\%) with standard derivation (std) on cancers data sets. The best results are in bold font.}}
\centering
\scriptsize
\begin{tabular}{|c|c c c c|c c c c| c c c c| }
\hline
Data sets        & \multicolumn{4}{c|}{Leukemia}        & \multicolumn{4}{c|}{Lung Cancer}    & \multicolumn{4}{c|}{Novartis BPLC}      \\
Side-info.       & 0\%     & 5\%     &   10\%  &15\% & 0\%      &5\%   & 10\% &15\% & 0\%            & 5\%      & 10\% &15\%      \\
\hline
\hline
LRR              &14.11 &    -     &    -    &  -      &5.08      &  -   &  -     &  -       &14.60      &  -        &  -    & - \\
%
LSR              &9.27  &    -     &    -    &  -      &4.57       &  -   &  -     &  -      &6.80       &  -        &  -    & - \\
PSC(1) \cite{McWilliams:DMKD14} &3.10  &    -      &    -    &  -      &7.80    &  -   &  -     &  -       &\text{4.60}  & -        &  -    & - \\
SSC              &3.23  &    -      &    -    &  -      &5.08    &  -   &  -     &  -      &\textbf{2.91}  & -        &  -    & - \\
\hline
\text{hard \cs3c} &\textbf{2.42} &\!\textbf{1.33$\pm$0.49} &\textbf{0.58$\pm$0.42}&\textbf{0.24$\pm$0.27}     &\textbf{4.06}    &\!\textbf{3.81$\pm$0.48}   &{\textbf{3.53$\pm$0.31}}&\textbf{2.94$\pm$0.58}          &\textbf{2.91}         &\!\textbf{2.33$\pm$1.02}   &\textbf{1.17$\pm$0.87} &\textbf{0.44$\pm$0.50} \\


\text{soft \cs3c} &\textbf{2.42} &\!\textbf{1.23$\pm$0.49} &\textbf{0.56$\pm$0.42}&\textbf{0.24$\pm$0.27} &\textbf{4.06} &\!\textbf{3.78$\pm$0.48} &{\textbf{3.50$\pm$0.31}}&\textbf{3.35$\pm$0.58} &\textbf{2.91}         &\!\textbf{2.28$\pm$1.02}    &{\textbf{0.97$\pm$0.87}}   &{\textbf{0.44$\pm$0.50}} \\

\hline
\end{tabular}
\label{table:cancers-SSC-vs-SSSC}
{\vspace{-5pt}}
\end{table*}

We consider the Hopkins 155 database \cite{Tron:CVPR07}, which consists of 155 video sequences
with 2 or 3 motions in each video corresponding to 2 or 3 low-dimensional subspaces. We compare our \s3c algorithms with SSC \cite{Elhamifar:TPAMI13}, LSA \cite{Yan:ECCV06}, LRR \cite{Liu:ICML10}, LSR \cite{Lu:ECCV12}, BDSSC \cite{Feng:CVPR14} and BDLRR \cite{Feng:CVPR14} on the Hopkins 155 motion segmentation data set \cite{Tron:CVPR07} for the multi-view affine motion segmentation without any other postprocessing (\eg, coefficients selection, thresholding, or $\ell_\infty$ normalization). 
{\mcb
In hard \s3c, $\alpha$ is set to $0.02$, which is tuned in the range $[0.01, 0.10]$; whereas in soft \s3c, $\alpha$ is set to $0.20$ which is tuned in the range $[0.02, 0.50]$.}
Experimental results are presented in Table~\ref{table:hopkins155-SSC-vs-SSSC}. 
our \s3c algorithms outperform SSC; however, due to the fact that the Hopkins 155 database has a relatively low noise level, the improvement in performance over SSC is relatively minor. Compared to hard \s3c, soft \s3c yields slightly better results because it captures more detailed information about the clustering results in the previous iterations by using continuous real-valued weights in $\Theta$. It is worth to note that there is no any postprocessing over the obtained coefficients matrix in our \s3c.
{\mcb 
The average time\footnote{MATLAB code on laptop (3.00GB RAM/
i5-2520M CPU@2.50GHz)} per sequence of SSC is 4.13s; whereas the average time in hard \s3c is 4.68s and 5.30s in soft \s3c. The average iteration number $\bar T_2$ of Algorithm \ref{alg:hard-S3C} in hard \s3c is 2.2; whereas $\bar T_2$ in soft \s3c is 2.5. We also observed that the iteration number $T_1$ in the first call of Algorithm \ref{alg:ST-SSR-ADMM} is between 100 and 150 and later is reduced to around 30. This accounts for the observation that the time cost per sequence in \s3c is not simply $T_2$ times that of SSC's.}

\subsection{Experiments on Cancer Gene Data Sets}

DNA microarray is a high-throughput technology that allows for simultaneously monitoring the mRNA levels of thousands of genes in particular cells or tissues \cite{Lockhart:Nature00, Schena:Science95, Schulze:NCB01}.
An important application of cancer gene expression data is to identify or discover subtypes of cancers \cite{Golub:Science99}.
Gene expression data 
from different subtypes of cancers lie on multiple clusters \cite{Alon:PNAS99}, each one 
relating to one subtype. As pointed out in \cite{McWilliams:DMKD14}, 
each cluster can be well approximated by a low-dimensional subspace.
If some genes expression profiles are known to have the same subtypes prior to performing an experiment, then this knowledge could be used 
as side-information. 
Thus, we incorporate the side-information for genes expression profiles clustering.



We consider three publicly available benchmark cancer data sets: 
St. Jude leukemia \cite{Yeoh:CC02-abbr}, Lung Cancer \cite{Bhattacharjee:PNAS01-abbr}, and Novartis BPLC \cite{Su:PNAS02-abbr}.
As baselines, we choose three popular spectral clustering based methods: SSC \cite{Elhamifar:TPAMI13}, LRR \cite{Liu:ICML10, Cui:PLOS13-LRR}, LSR \cite{Lu:ECCV12}, and a PCA based subspace clustering method, Predictive Subspace Clustering (PSC) \cite{McWilliams:DMKD14}. The results of PSC are directly cite from \cite{McWilliams:DMKD14}.
{\mcb For SSC, LRR, and LSR, we tune parameter $\lambda$ on each data set and record the best accuracy. Specifically, we use $\lambda_0=4, 10, 5$ for SSC, $\lambda=1.4, 0.5, 1$ for LRR, $\lambda=0.15, 5, 1$ for LSR, on the three data sets, respectively. For \cs3c, $\lambda$ is set to be the same as that in SSC and parameter $\alpha$ is tuned in the range $[10^{-4},10^{-1}]$ 
on each data set.} 
To prepare the side-information, we randomly sample a percentage ($p=5\%, 10\%, 15\%$) of entries from the ground truth structure matrix $\Theta_0$ whose entries encode the clustering membership. 
In \cs3c, we encode the side-information into the matrix $\Psi$. We record the average clustering error (ERR) over 20 trials. Experimental results are presented in Table \ref{table:cancers-SSC-vs-SSSC}. Note that \cs3c reduces to \s3c if there is not side-information available. 
It can be observed that: a) with the help of side-information, \cs3c can reduce the clustering errors notably; b) \cs3c without side-information (\ie, side-info. $0\%$), which is essentially \s3c, also matches or slightly outperforms SSC, LRR, and LSR. 

\section{Conclusion}
\label{sec:conclusion}
We proposed a joint optimization framework for the problem of subspace clustering, called Structured Sparse Subspace Clustering (\s3c), in which the separate two stages of computing the sparse representation and applying the spectral clustering are 
elegantly combined. 
In addition, we extended the \s3c framework into Constrained Structured Sparse Subspace Clustering (\cs3c) in which available partial side-information is incorporated 
into the stage of learning the affinity. We solved the optimization problem via efficient alternating minimization algorithm that combines an alternating direction method of multipliers and spectral clustering.
Experiments on a synthetic data set, the Extended Yale B face data set, the Hopkins 155 motion segmentation database, and three cancer gene data sets demonstrated the effectiveness of our approach.
{\mcb Finally, we note that the proposed joint optimization framework could be extended to 
address the problem of multi-view clustering \cite{Yao:arXiv16-S3CAM} and semi-supervised learning \cite{Li:ICCV15-STSSL}. Besides, one can also explore using more advanced spectral clustering techniques, \eg, \cite{Lu:TIP16}. The scalability and theoretical justifications for the joint optimization framework are also interesting future work.

}

\section*{Acknowledgment}
C.-G. Li was partially supported by the National Natural Science Foundation of China under Grant No. 61273217, 
the Scientific Research Foundation for the Returned Overseas Chinese Scholars, Ministry of Education of China,
and the Open Project Fund from MOE Key Laboratory of Machine Perception, Peking University. C. You and R. Vidal were supported by the National Science Foundation under Grant No. 1447822.

{\small
\bibliographystyle{IEEEtran}
\bibliography{IEEEabrv,biblio/temp,biblio/cgli,biblio/vidal,biblio/vision,biblio/math,biblio/learning,biblio/sparse,biblio/geometry,biblio/dti,biblio/recognition,biblio/surgery,biblio/coding,biblio/matrixcompletion,biblio/segmentation,biblio/computationalbiology}

\begin{thebibliography}{100}
\providecommand{\url}[1]{#1}
\csname url@samestyle\endcsname
\providecommand{\newblock}{\relax}
\providecommand{\bibinfo}[2]{#2}
\providecommand{\BIBentrySTDinterwordspacing}{\spaceskip=0pt\relax}
\providecommand{\BIBentryALTinterwordstretchfactor}{4}
\providecommand{\BIBentryALTinterwordspacing}{\spaceskip=\fontdimen2\font plus
\BIBentryALTinterwordstretchfactor\fontdimen3\font minus
  \fontdimen4\font\relax}
\providecommand{\BIBforeignlanguage}[2]{{%
\expandafter\ifx\csname l@#1\endcsname\relax
\typeout{** WARNING: IEEEtran.bst: No hyphenation pattern has been}%
\typeout{** loaded for the language `#1'. Using the pattern for}%
\typeout{** the default language instead.}%
\else
\language=\csname l@#1\endcsname
\fi
#2}}
\providecommand{\BIBdecl}{\relax}
\BIBdecl

\bibitem{Tomasi:IJCV92}
C.~Tomasi and T.~Kanade, ``Shape and motion from image streams under
  orthography,'' \emph{International Journal of Computer Vision}, vol.~9,
  no.~2, pp. 137--154, 1992.

\bibitem{Ho:CVPR03}
J.~Ho, M.~H. Yang, J.~Lim, K.~Lee, and D.~Kriegman, ``Clustering appearances of
  objects under varying illumination conditions.'' in \emph{{IEEE} Conference
  on Computer Vision and Pattern Recognition}, 2003, pp. 11--18.

\bibitem{Hong:TIP06}
W.~Hong, J.~Wright, K.~Huang, and Y.~Ma, ``Multi-scale hybrid linear models for
  lossy image representation,'' \emph{{IEEE} Transactions on Image Processing},
  vol.~15, no.~12, pp. 3655--3671, 2006.

\bibitem{Costeira:IJCV98}
J.~Costeira and T.~Kanade, ``A multibody factorization method for independently
  moving objects,'' \emph{International Journal of Computer Vision}, vol.~29,
  no.~3, pp. 159--179, 1998.

\bibitem{Rao:PAMI10}
S.~Rao, R.~Tron, R.~Vidal, and Y.~Ma, ``Motion segmentation in the presence of
  outlying, incomplete, or corrupted trajectories,'' \emph{{IEEE} Transactions
  on Pattern Analysis and Machine Intelligence}, vol.~32, no.~10, pp.
  1832--1845, 2010.

\bibitem{Vidal:PAMI05}
R.~Vidal, Y.~Ma, and S.~Sastry, ``{Generalized Principal Component Analysis
  (GPCA)},'' \emph{{IEEE} Transactions on Pattern Analysis and Machine
  Intelligence}, vol.~27, no.~12, pp. 1--15, 2005.

\bibitem{Bako:Automatica11}
L.~Bako, ``Identification of switched linear systems via sparse optimization,''
  \emph{Automatica}, vol.~47, no.~4, pp. 668--677, 2011.

\bibitem{Jalali:ICML11}
A.~Jalali, Y.~Chen, S.~Sanghavi, and H.~Xu, ``Clustering partially observed
  graphs via convex optimization,'' in \emph{International Conference on
  Machine Learning}, no. 1001-1008, 2011.

\bibitem{McWilliams:DMKD14}
B.~McWilliams and G.~Montana, ``Subspace clustering of high dimensional data: a
  predictive approach,'' \emph{Data Mining and Knowledge Discovery}, vol.~28,
  no.~3, pp. 736--772, 2014.

\bibitem{Bradley:JGO00}
P.~S. Bradley and O.~L. Mangasarian, ``k-plane clustering,'' \emph{Journal of
  Global Optimization}, vol.~16, no.~1, pp. 23--32, 2000.

\bibitem{Tseng:JOTA00}
P.~Tseng, ``Nearest $q$-flat to $m$ points,'' \emph{Journal of Optimization
  Theory and Applications}, vol. 105, no.~1, pp. 249--252, 2000.

\bibitem{Zhang:WSM09}
T.~Zhang, A.~Szlam, and G.~Lerman, ``Median $k$-flats for hybrid linear
  modeling with many outliers,'' in \emph{Workshop on Subspace Methods}, 2009,
  pp. 234--241.

\bibitem{Agarwal:ACM04}
P.~Agarwal and N.~Mustafa, ``k-means projective clustering,'' in \emph{ACM
  Symposium on Principles of database systems}, 2004.

\bibitem{Ma:SIAM08}
Y.~Ma, A.~Y. Yang, H.~Derksen, and R.~Fossum, ``Estimation of subspace
  arrangements with applications in modeling and segmenting mixed data,''
  \emph{SIAM Review}, vol.~50, no.~3, pp. 413--458, 2008.

\bibitem{Huang:CVPR04-ED}
K.~Huang, Y.~Ma, and R.~Vidal, ``Minimum effective dimension for mixtures of
  subspaces: A robust {GPCA}, algorithm and its applications,'' in \emph{{IEEE}
  Conference on Computer Vision and Pattern Recognition}, vol.~II, 2004, pp.
  631--638.

\bibitem{Tsakiris:AffineASC-ArXiv15}
M.~C. Tsakiris and R.~Vidal, ``Algebraic clustering of affine subspaces,''
  \emph{ArXiv}, 2015.

\bibitem{Boult:WMU91}
T.~Boult and L.~Brown, ``Factorization-based segmentation of motions,'' in
  \emph{IEEE Workshop on Motion Understanding}, 1991, pp. 179--186.

\bibitem{Leonardis:PR02}
A.~Leonardis, H.~Bischof, and J.~Maver, ``Multiple eigenspaces,'' \emph{Pattern
  Recognition}, vol.~35, no.~11, pp. 2613--2627, 2002.

\bibitem{Archambeau:Neuro08}
C.~Archambeau, N.~Delannay, and M.~Verleysen, ``Mixtures of robust
  probabilistic principal component analyzers,'' \emph{Neurocomputing},
  vol.~71, no. 7--9, pp. 1274--1282, 2008.

\bibitem{Gruber-Weiss:CVPR04}
A.~Gruber and Y.~Weiss, ``Multibody factorization with uncertainty and missing
  data using the {EM} algorithm,'' in \emph{IEEE Conference on Computer Vision
  and Pattern Recognition}, vol.~I, 2004, pp. 707--714.

\bibitem{Ma:PAMI07}
Y.~Ma, H.~Derksen, W.~Hong, and J.~Wright, ``Segmentation of multivariate mixed
  data via lossy coding and compression,'' \emph{IEEE Transactions on Pattern
  Analysis and Machine Intelligence}, vol.~29, no.~9, pp. 1546--1562, 2007.

\bibitem{Yang:CVPR06}
A.~Y. Yang, S.~Rao, and Y.~Ma, ``Robust statistical estimation and segmentation
  of multiple subspaces,'' in \emph{Workshop on 25 years of RANSAC}, 2006.

\bibitem{Yan:ECCV06}
J.~Yan and M.~Pollefeys, ``A general framework for motion segmentation:
  Independent, articulated, rigid, non-rigid, degenerate and non-degenerate,''
  in \emph{European Conference on Computer Vision}, 2006, pp. 94--106.

\bibitem{Goh:CVPR07}
A.~Goh and R.~Vidal, ``Segmenting motions of different types by unsupervised
  manifold clustering,'' in \emph{{IEEE} Conference on Computer Vision and
  Pattern Recognition}, 2007, pp. 1--6.

\bibitem{Fan:PAMI06}
Z.~Fan, J.~Zhou, and Y.~Wu, ``Multibody grouping by inference of multiple
  subspaces from high-dimensional data using oriented-frames,'' \emph{{IEEE}
  Transactions on Pattern Analysis and Machine Intelligence}, vol.~28, no.~1,
  pp. 91--105, 2006.

\bibitem{Chen:IJCV09}
G.~Chen and G.~Lerman, ``Spectral curvature clustering ({SCC}),''
  \emph{International Journal of Computer Vision}, vol.~81, no.~3, pp.
  317--330, 2009.

\bibitem{Zhang:IJCV12}
T.~Zhang, A.~Szlam, Y.~Wang, and G.~Lerman, ``Hybrid linear modeling via local
  best-fit flats,'' \emph{International Journal of Computer Vision}, vol. 100,
  no.~3, pp. 217--240, 2012.

\bibitem{Elhamifar:CVPR09}
E.~Elhamifar and R.~Vidal, ``Sparse subspace clustering,'' in \emph{{IEEE}
  Conference on Computer Vision and Pattern Recognition}, 2009, pp. 2790--2797.

\bibitem{Elhamifar:TPAMI13}
------, ``Sparse subspace clustering: Algorithm, theory, and applications,''
  \emph{{IEEE} Transactions on Pattern Analysis and Machine Intelligence},
  vol.~35, no.~11, pp. 2765--2781, 2013.

\bibitem{Liu:ICML10}
G.~Liu, Z.~Lin, and Y.~Yu, ``Robust subspace segmentation by low-rank
  representation,'' in \emph{International Conference on Machine Learning},
  2010, pp. 663--670.

\bibitem{Liu:TPAMI13}
G.~Liu, Z.~Lin, S.~Yan, J.~Sun, and Y.~Ma, ``Robust recovery of subspace
  structures by low-rank representation,'' \emph{IEEE Transactions on Pattern
  Analysis and Machine Intelligence}, vol.~35, no.~1, pp. 171--184, Jan 2013.

\bibitem{Favaro:CVPR11}
P.~Favaro, R.~Vidal, and A.~Ravichandran, ``A closed form solution to robust
  subspace estimation and clustering,'' in \emph{IEEE Conference on Computer
  Vision and Pattern Recognition}, 2011, pp. 1801 --1807.

\bibitem{Vidal:PRL14}
R.~Vidal and P.~Favaro, ``Low rank subspace clustering {(LRSC)},''
  \emph{Pattern Recognition Letters}, vol.~43, pp. 47--61, 2014.

\bibitem{Lu:ECCV12}
C.-Y. Lu, H.~Min, Z.-Q. Zhao, L.~Zhu, D.-S. Huang, and S.~Yan, ``Robust and
  efficient subspace segmentation via least squares regression,'' in
  \emph{European Conference on Computer Vision}, 2012, pp. 347--360.

\bibitem{Lu:ICCV13-TraceLasso}
C.~Lu, Z.~Lin, and S.~Yan, ``Correlation adaptive subspace segmentation by
  trace lasso,'' in \emph{{IEEE} International Conference on Computer Vision},
  2013, pp. 1345--1352.

\bibitem{Lu:ICCV13}
C.~Lu, S.~Yan, and Z.~Lin, ``Correntropy induced l2 graph for robust subspace
  clustering,'' in \emph{{IEEE} International Conference on Computer Vision},
  2013, pp. 1801--1808.

\bibitem{Wang:NIPS13-LRR+SSC}
Y.-X. Wang, H.~Xu, and C.~Leng, ``Provable subspace clustering: When {LRR}
  meets {SSC},'' in \emph{Neural Information Processing Systems}, 2013.

\bibitem{Dyer:JMLR13}
E.~L. Dyer, A.~C. Sankaranarayanan, and R.~G. Baraniuk, ``Greedy feature
  selection for subspace clustering,'' \emph{Journal of Machine Learning
  Research}, vol.~14, no.~1, pp. 2487--2517, 2013.

\bibitem{Heckel:TIT15}
R.~Heckel and H.~B{\"o}lcskei, ``Robust subspace clustering via thresholding,''
  \emph{IEEE Transactions on Information Theory}, vol.~61, no.~11, pp.
  6320--6342, 2015.

\bibitem{Zhang:ICCV13}
Y.~Zhang, Z.~Sun, R.~He, and T.~Tan, ``Robust subspace clustering via
  half-quadratic minimization,'' in \emph{{IEEE} International Conference on
  Computer Vision}, 2013, pp. 3096--3103.

\bibitem{Park:NIPS14}
D.~Park, C.~Caramanis, and S.~Sanghavi, ``Greedy subspace clustering,'' in
  \emph{Neural Information Processing Systems}, 2014.

\bibitem{Li:CVPR15MoG}
B.~Li, Y.~Zhang, Z.~Lin, and H.~Lu, ``Subspace clustering by mixture of
  gaussian regression,'' in \emph{{IEEE} Conference on Computer Vision and
  Pattern Recognition}, 2015, pp. 2094--2102.

\bibitem{You:CVPR16-EnSC}
C.~You, C.-G. Li, D.~Robinson, and R.~Vidal, ``Oracle based active set
  algorithm for scalable elastic net subspace clustering,'' in
  \emph{Proceedings of {IEEE} International Conference on Computer Vision and
  Pattern Recognition}, 2016, pp. 3928--3937.

\bibitem{Vidal:SPM11-SC}
R.~Vidal, ``Subspace clustering,'' \emph{{IEEE} Signal Processing Magazine},
  vol.~28, no.~3, pp. 52--68, March 2011.

\bibitem{Fazel2002}
M.~Fazel, ``Matrix rank minimization with applications,'' Ph.D. dissertation,
  Stanford University, 2002.

\bibitem{Grave:NIPS11}
E.~Grave, G.~Obozinski, and F.~Bach, ``Trace lasso: a trace norm regularization
  for correlated designs,'' in \emph{Neural Information Processing Systems},
  2011.

\bibitem{Zhang:NeuralCom15}
H.~Zhang, Z.~Lin, C.~Zhang, and J.~Gao, ``Relation among some low rank subspace
  recovery models,'' \emph{Neural Computation}, vol.~27, no.~9, pp. 1915--1950,
  2015.

\bibitem{Peng:TNNLS16}
X.~Peng, C.~Lu, Z.~Yi, and H.~Tang, ``Connections between nuclear-norm and
  frobenius-norm-based representations,'' \emph{IEEE Transactions on Neural
  Networks and Learning Systems}, vol. 99, to appear, 2016.

\bibitem{Elhamifar:ICASSP10}
E.~Elhamifar and R.~Vidal, ``Clustering disjoint subspaces via sparse
  representation,'' in \emph{{IEEE} International Conference on Acoustics,
  Speech, and Signal Processing}, 2010, pp. 1926--1929.

\bibitem{You:CVPR16-SSCOMP}
C.~You, D.~Robinson, and R.~Vidal, ``Scalable sparse subspace clustering by
  orthogonal matching pursuit,'' in \emph{{IEEE} Conference on Computer Vision
  and Pattern Recognition}, 2016, pp. 3918--3927.

\bibitem{Luo:ECML11}
D.~Luo, F.~Nie, C.~H.~Q. Ding, and H.~Huang, ``Multi-subspace representation
  and discovery,'' in \emph{ECML/PKDD}, 2011, pp. 405--420.

\bibitem{Li:ACCV09}
C.-G. Li, J.~Guo, and H.~Zhang, ``Learning bundle manifold by double
  neighborhood graphs,'' in \emph{Asian Conference on Computer Vision, Part
  III, LNCS}, vol. 5996, 2009, pp. 321--330.

\bibitem{Elhamifar:NIPS11}
E.~Elhamifar and R.~Vidal, ``Sparse manifold clustering and embedding,'' in
  \emph{Neural Information Processing and Systems}, 2011.

\bibitem{Peng:TCYB16}
X.~Peng, Z.~Yu, Z.~Yi, and H.~Tang, ``Constructing the $l_2$-graph for robust
  subspace learning and subspace clustering,'' \emph{IEEE Transactions on
  Cybernetics}, vol. 99, to appear, 2016.

\bibitem{Tschannen:arXiv16}
M.~Tschannen and H.~Bolcskei, ``Noisy subspace clustering via matching
  pursuits,'' \emph{arXiv:1612.03450}, 2016.

\bibitem{Wang:AISTAT16}
Y.~Wang, Y.-X. Wang, and A.~Singh, ``Graph connectivity in noisy sparse
  subspace clustering,'' in \emph{Proceedings of the 19th International
  Conference on Artificial Intelligence and Statistics}, 2016, pp. 538--546.

\bibitem{Yang:ECCV16}
Y.~Yang, J.~Feng, N.~Jojic, J.~Yang, and T.~S. Huang, ``$\ell_0$-sparse
  subspace clustering,'' in \emph{European Conference on Computer Vision},
  2016, pp. 731--747.

\bibitem{Pham:CVPR12}
D.~Pham, S.~Budhaditya, D.~Phung, and S.~Venkatesh, ``Improved subspace
  clustering via exploitation of spatial constraints,'' in \emph{{IEEE}
  Conference on Computer Vision and Pattern Recognition}, 2012, pp. 550--557.

\bibitem{Hu:CVPR14}
H.~Hu, Z.~Lin, J.~Feng, and J.~Zhou, ``Smooth representation clustering,'' in
  \emph{{IEEE} Conference on Computer Vision and Pattern Recognition}, 2014,
  pp. 3834--3841.

\bibitem{Patel:ICCV13}
V.~M. Patel, H.~V. Nguyen, and R.~Vidal, ``Latent space sparse subspace
  clustering,'' in \emph{{IEEE} International Conference on Computer Vision},
  2013, pp. 225--232.

\bibitem{Patel:ICIP14}
V.~M. Patel and R.~Vidal, ``Kernel sparse subspace clustering,'' in
  \emph{{IEEE} International Conference on Image Processing}, 2014, pp.
  2849--2853.

\bibitem{Patel:JSTSP15}
V.~M. Patel, H.~V. Nguyen, and R.~Vidal, ``Latent space sparse and low-rank
  subspace clustering,'' \emph{{IEEE} Journal of Selected Topics in Signal
  Processing}, vol.~9, no.~4, pp. 691--701, 2015.

\bibitem{Shen:ICML16}
J.~Shen, P.~Li, and H.~Xu, ``Online low-rank subspace clustering by basis
  dictionary pursuit,'' in \emph{Proceedings of the 33rd International
  Conference on Machine Learning}, 2016, pp. 622--631.

\bibitem{vonLuxburg:StatComp2007}
U.~von Luxburg, ``A tutorial on spectral clustering,'' \emph{Statistics and
  Computing}, vol.~17, no.~4, pp. 395--416, 2007.

\bibitem{Li:CVPR15}
C.-G. Li and R.~Vidal, ``Structured sparse subspace clustering: A unified
  optimization framework,'' in \emph{Proceedings of {IEEE} International
  Conference on Computer Vision and Pattern Recognition}, 2015, pp. 277--286.

\bibitem{Vidal:Springer16}
R.~Vidal, Y.~Ma, and S.~Sastry, \emph{Generalized Principal Component
  Analysis}.\hskip 1em plus 0.5em minus 0.4em\relax Springer Verlag, 2016.

\bibitem{Boyd:FTML10}
S.~Boyd, N.~Parikh, E.~Chu, B.~Peleato, and J.~Eckstein, ``Distributed
  optimization and statistical learning via the alternating direction method of
  multipliers,'' \emph{Foundations and Trends in Machine Learning}, vol.~3,
  no.~1, pp. 1--122, 2010.

\bibitem{Lin:09}
Z.~Lin, M.~Chen, L.~Wu, and Y.~Ma, ``The augmented {Lagrange} multiplier method
  for exact recovery of corrupted low-rank matrices,''
  \emph{arXiv:1009.5055v2}, 2011.

\bibitem{Li:SPARS15}
C.-G. Li, C.~You, and R.~Vidal, ``On sufficient conditions for affine sparse
  subspace clustering,'' in \emph{Signal Processing with Adaptive Sparse
  Structured Representations}, 2015.

\bibitem{Soltanolkotabi:AS12}
M.~Soltanolkotabi and E.~J. Cand\`es, ``A geometric analysis of subspace
  clustering with outliers,'' \emph{Annals of Statistics}, vol.~40, no.~4, pp.
  2195--2238, 2012.

\bibitem{Wang-Xu:ICML13}
Y.-X. Wang and H.~Xu, ``Noisy sparse subspace clustering,'' in
  \emph{International Conference on Machine Learning}, 2013, pp. 89--97.

\bibitem{Soltanolkotabi:AS14}
M.~Soltanolkotabi, E.~Elhamifar, and E.~J. Cand\`es, ``Robust subspace
  clustering,'' \emph{Annals of Statistics}, vol.~42, no.~2, pp. 669--699,
  2014.

\bibitem{Wang:JMLR16}
Y.-X. Wang and H.~Xu, ``Noisy sparse subspace clustering,'' \emph{Journal of
  Machine Learning Research}, vol.~17, no.~12, pp. 1--41, 2016.

\bibitem{Wang:ICML15}
Y.~Wang, Y.~Wang, and A.~Singh, ``A deterministic analysis of noisy sparse
  subspace clustering for dimensionality-reduced data,'' in \emph{International
  Conference on Machine Learning}, 2015, pp. 1422--1431.

\bibitem{Candes:JFAA08}
E.~Cand\`es, M.~Wakin, and S.~Boyd, ``Enhancing sparsity by reweighted $\ell_1$
  minimization,'' \emph{Journal of Fourier Analysis and Applications}, vol.~14,
  no.~5, pp. 877--905, 2008.

\bibitem{Feng:CVPR14}
J.~Feng, Z.~Lin, H.~Xu, and S.~Yan, ``Robust subspace segmentation with
  block-diagonal prior,'' in \emph{{IEEE} Conference on Computer Vision and
  Pattern Recognition}, 2014, pp. 3818--3825.

\bibitem{Guo:IJCAI15}
X.~Guo, ``Robust subspace segmentation by simultaneously learning data
  representations and their affinity matrix,'' in \emph{Proceedings of the 24th
  International Joint Conference on Artificial Intelligence}, 2015, pp.
  3547--3553.

\bibitem{Peng:CVPR13}
X.~Peng, L.~Zhang, and Z.~Yi, ``Scalable sparse subspace clustering,''
  \emph{{IEEE} Conference on Computer Vision and Pattern Recognition}, pp.
  430--437, 2013.

\bibitem{Kodirov:ECCV16}
E.~Kodirov, T.~Xiang, Z.~Fu, and S.~Gong, ``Person re-identification by
  unsupervised $\ell_1$ graph learning,'' in \emph{European Conference on
  Computer Vision}, 2016, pp. 178--195.

\bibitem{Li:TSP16}
C.-G. Li and R.~Vidal, ``A structured sparse plus structured low-rank framework
  for subspace clustering and completion,'' \emph{{IEEE} Transactions on Signal
  Processing}, vol.~64, no.~24, pp. 6557--6570, 2016.

\bibitem{Fang:JBI06}
Z.~Fang, J.~Yang, Y.~Li, Q.~Luo, L.~Liu, and et~al., ``Knowledge guided
  analysis of microarray data,'' \emph{Journal of Biomedical Informatics},
  vol.~39, no.~4, pp. 401--411, 2006.

\bibitem{Chopra:BMCbioinfo08}
P.~Chopra, J.~Kang, J.~Yang, H.~Cho, H.~Kim, and M.~Lee, ``Microarray data
  mining using landmark gene-guided clustering,'' \emph{BMC Bioinformatics},
  vol.~9, no.~1, p.~92, 2008.

\bibitem{Huang:Bioinfo06}
D.~Huang and W.~Pan, ``Incorporating biological knowledge into distance-based
  clustering analysis of microarray gene expression data,''
  \emph{Bioinformatics}, vol.~22, no.~10, pp. 1259--1268, 2006.

\bibitem{Bair:WIRCS13}
E.~Bair, ``Semi-supervised clustering methods,'' \emph{Wiley Interdisciplinary
  Reviews: Computational Statistics}, vol.~5, no.~5, pp. 349--361, 2013.

\bibitem{Shi-Malik:PAMI00}
J.~Shi and J.~Malik, ``Normalized cuts and image segmentation,'' \emph{{IEEE}
  Transactions on Pattern Analysis and Machine Intelligence}, vol.~22, no.~8,
  pp. 888--905, 2000.

\bibitem{Fiedler:CMathJ1975}
M.~Fiedler, ``\BIBforeignlanguage{English}{{A property of eigenvectors of
  nonnegative symmetric matrices and its application to graph theory.}}''
  \emph{\BIBforeignlanguage{English}{{Czech. Math. J.}}}, vol.~25, pp.
  619--633, 1975.

\bibitem{Kriegman:PAMI01}
A.~Georghiades, P.~Belhumeur, and D.~Kriegman, ``From few to many: Illumination
  cone models for face recognition under variable lighting and pose,''
  \emph{IEEE Transactions on Pattern Analysis and Machine Intelligence},
  vol.~23, no.~6, pp. 643--660, 2001.

\bibitem{Liu:ICCV11}
G.~Liu and S.~Yan, ``Latent low-rank representation for subspace segmentation
  and feature extraction,'' in \emph{{IEEE} International Conference on
  Computer Vision}, 2011, pp. 1615--1622.

\bibitem{Tron:CVPR07}
R.~Tron and R.~Vidal, ``A benchmark for the comparison of 3-{D} motion
  segmentation algorithms,'' in \emph{{IEEE} Conference on Computer Vision and
  Pattern Recognition}, 2007, pp. 1--8.

\bibitem{Lockhart:Nature00}
D.~Lockhart and E.~Winzeler, ``Genomics, gene expression, and dna arrays,''
  \emph{Nature}, vol. 405, pp. 827--836, 2000.

\bibitem{Schena:Science95}
M.~Schena, D.~Shalon, R.~Davis, and P.~Brown, ``Quantitative monitoring of gene
  expression patterns with a complementary dna microarray,'' \emph{Science},
  vol. 270, pp. 467--470, 1995.

\bibitem{Schulze:NCB01}
A.~Schulze and J.~Downward, ``Navigating gene expression using microarrays -- a
  technology review,'' \emph{Nat. Cell Biol.}, vol.~3, pp. E190--E195, 2001.

\bibitem{Golub:Science99}
T.~Golub, D.~Slonim, P.~Tamayo, C.~Huard, M.~Gaasenbeek, J.~Mesirov, H.~Coller,
  M.~Loh, J.~Downing, M.~Caligiuri, C.~Bloomfield, and E.~Lander, ``Molecular
  classification of cancer: class discovery and class prediction by genes
  expression monitoring,'' \emph{Science}, vol. 286, no. 5439, pp. 531--537,
  1999.

\bibitem{Alon:PNAS99}
U.~Alon, N.~Barkai, D.~Notterman, K.~Gish, S.~Ybarra, D.~Mack, and A.~Levine,
  ``Broad patterns of gene expression revealed by clustering analysis of tumor
  and normal colon tissues probed by oligonucleotide arrays.''
  \emph{Proceedings of the National Academy of Sciences of the United States of
  America}, vol.~96, no.~12, pp. 6745--6750, 1999.

\bibitem{Yeoh:CC02-abbr}
E.~J. Yeoh, M.~E. Ross, S.~A. Shurtleff, and et~al., ``Classification, subtype
  discovery, and prediction of outcome in pediatric acute lymphoblastic
  leukemia by gene expression profiling,'' \emph{Cancer Cell}, vol.~1, pp.
  133--143, 2002.

\bibitem{Bhattacharjee:PNAS01-abbr}
A.~Bhattacharjee, W.~Richards, J.~Staunton, and et~al., ``Classification of
  human lung carcinomas by mrna expression profiling reveals distinct
  adenocarcinomas sub-classes,'' \emph{Proceedings of the National Academy of
  Sciences of the United States of America}, vol.~98, no.~24, pp.
  13,790--13,795, 2001.

\bibitem{Su:PNAS02-abbr}
A.~I. Su, M.~P. Cooke, K.~A. Ching, and et~al., ``Large-scale analysis of the
  human and mouse transcriptomes,'' \emph{Proceedings of the National Academy
  of Sciences of the United States of America}, vol.~99, no.~7, pp. 4447--4465,
  2002.

\bibitem{Cui:PLOS13-LRR}
Y.~Cui, C.-H. Zheng, and J.~Yang, ``Identifying subspace gene clusters from
  microarray data using low-rank representation,'' \emph{Plos ONE}, vol.~8,
  no.~3, p. e59377, 2013.

\bibitem{Yao:arXiv16-S3CAM}
J.~Yao and F.~Nielsen, ``{SSSC-AM}: A unified framework for video
  co-segmentation by structured sparse subspace clustering with appearance and
  motion features,'' \emph{arXiv:1603.04139}, 2016.

\bibitem{Li:ICCV15-STSSL}
C.-G. Li, Z.~Lin, H.~Zhang, and J.~Guo, ``Learning semi-supervised
  representation towards a unified optimization framework for semi-supervised
  learning,'' in \emph{Proceedings of {IEEE} International Conference on
  Computer Vision}, 2015, pp. 2767--2775.

\bibitem{Lu:TIP16}
C.~Lu, S.~Yan, and Z.~Lin, ``Convex sparse spectral clustering: Single-view to
  multi-view,'' \emph{IEEE Trans. Image Processing}, vol.~25, no.~6, pp.
  2833--2843, 2016.

\end{thebibliography}
}

\end{document}